\pdfoutput=1
\PassOptionsToPackage{table}{xcolor}
\documentclass[11pt]{article}
\usepackage[final]{ACL2023}
\usepackage{times}
\usepackage{latexsym}
\usepackage[T1]{fontenc}
\usepackage[utf8]{inputenc}
\usepackage{placeins}
\usepackage{microtype}
\usepackage{inconsolata}
\usepackage{amssymb}
\usepackage{multirow}
\usepackage{graphicx}
\usepackage{float}
\usepackage[table]{xcolor}
\usepackage{tcolorbox}
\definecolor{black}{HTML}{000000}
\definecolor{white}{HTML}{FFFFFF}
\definecolor{blue}{HTML}{AADDFF}
\definecolor{green}{HTML}{AAFFAA}
\definecolor{yellow}{HTML}{FFFF88}
\definecolor{red}{HTML}{FF8888}
\definecolor{orange}{HTML}{FF7700}
\definecolor{purple}{HTML}{AAAAFF}
\definecolor{pink}{HTML}{FFAAFF}
\definecolor{grey}{HTML}{BBBBBB}
\definecolor{color1}{HTML}{00AA00}
\definecolor{color2}{HTML}{AA0000}
\definecolor{color3}{HTML}{0000FF}
\title{DSBC : Data Science task Benchmarking with Context engineering}
\author{
 \textbf{Ram Mohan Rao Kadiyala \textsuperscript{1,2}},
 \textbf{Siddhant Gupta \textsuperscript{2,3}},
 \textbf{Jebish Purbey \textsuperscript{2}},\\
 \textbf{Giulio Martini \textsuperscript{1}},
 \textbf{Ali Shafique \textsuperscript{1}}
 \textbf{Suman Debnath  \textsuperscript{4}},
 \textbf{Hamza Farooq           \textsuperscript{1,5}}\\
 \\
 \textsuperscript{1}Traversaal.ai, 
 \textsuperscript{2}Cohere Labs Community,\\
 \textsuperscript{3}IIT Roorkee,
 \textsuperscript{4}Amazon,
 \textsuperscript{5}UCLA,\\
 \\
 \small{
   \textbf{Correspondence:} \href{mailto:contact@rkadiyala.com}{ram@traversaal.ai}
 }
}
\begin{document}
\maketitle
\begin{abstract}
Recent advances in large language models (LLMs) have significantly impacted data science workflows, giving rise to specialized data science agents designed to automate analytical tasks. Despite rapid adoption, systematic benchmarks evaluating the efficacy and limitations of these agents remain scarce. In this paper, we introduce a comprehensive benchmark specifically crafted to reflect real-world user interactions with data science agents by observing usage of our commercial applications. We evaluate three LLMs: Claude-4.0-Sonnet, Gemini-2.5-Flash, and OpenAI-o4-Mini across three approaches: zero-shot with context engineering, multi-step with context engineering, and with SmolAgent. Our benchmark assesses performance across a diverse set of eight data science task categories, additionally exploring the sensitivity of models to common prompting issues, such as data leakage and slightly ambiguous instructions. We further investigate the influence of temperature parameters on overall and task-specific outcomes for each model and approach. Our findings reveal distinct performance disparities among the evaluated models and methodologies, highlighting critical factors that affect practical deployment. The benchmark dataset and evaluation framework introduced herein aim to provide a foundation for future research of more robust and effective data science agents.
\end{abstract}
\section{Introduction}
\label{sec:introduction}
Large Language Models (LLMs) have recently gained prominence due to their capability to automate and enhance various data science tasks. This growing capability has led to increased adoption of specialized data science agents across multiple domains. Despite widespread usage, there is a clear gap in comprehensive evaluations that accurately reflect practical user interactions and realistic task scenarios. This gap makes it challenging for practitioners and researchers to understand the true efficacy and limitations of these agents in applied settings.

In response to this, we introduce a detailed benchmark tailored to reflect actual usage patterns of data science agents, derived from observations of end-user behavior. We evaluate three leading LLMs, Claude-4.0-Sonnet \citep{claude4}, Gemini 2.5 Flash \citep{comanici2025gemini25pushingfrontier}, and OpenAI-o4-Mini \citep{openai_o4mini_2025}, using three distinct methodologies: zero-shot with context engineering, multi-step with context engineering, and using SmolAgent \citep{smolagents}. Our benchmark covers diverse and practical data science tasks while also examining how sensitive these models are to common prompting issues like data leakage and slight ambiguity.

The performance of agents or Large Language Models (LLMs) critically depends on the context provided during inference, including both the maximum context length available and the efficiency of its utilization. \citep{mei2025surveycontextengineeringlarge}. While most benchmarks come with a manually written or synthetically generated summary or a description of the datasets used, we use a standardized context through context engineering.

Additionally, we analyze the impact of varying temperature settings on both overall performance and task-specific effectiveness. Our findings underscore significant differences in the capabilities of evaluated models and strategies, identifying crucial considerations for practical deployment. Through this work, we aim to provide a foundational resource for furthering the development of more reliable and effective data science agents.
\begin{table*}[!t]
    \centering
    \resizebox{\textwidth}{!}{\
    \begin{tabular}{ccccccccc}
    \noalign{\hrule height 1pt}
    \rowcolor{orange}\textbf{\small{Prior Work}} & \textbf{\small{Domain of}} & \textbf{\small{Context added}} & \textbf{\small{Avg. row count}} & \textbf{\small{Avg. col count}} & \textbf{\small{Span Multiple}} & \textbf{\small{Prompt}} & \textbf{\small{Temp.}} & \textbf{\small{Sample}} \\
    \rowcolor{orange}\textbf{\small{}} & \textbf{\small{Benchmark}} & \textbf{\small{about data files}} & \textbf{\small{In Data files}} & \textbf{\small{In Data files}} & \textbf{\small{task types}} & \textbf{\small{Variations}} & \textbf{\small{Variations}} & \textbf{\small{Count}} \\
    \noalign{\hrule height 1pt}
    
    \noalign{\hrule height 1pt}
    \multicolumn{9}{l}{\small{\textbf{Other Domain}}}\\
    \noalign{\hrule height 1pt}
    \small{Spider \citep{yu-etal-2018-spider}}    & \small{Text-to-SQL}          & N/A & - & - & - & No & No & 1,034 \\
    \small{MLAgentBench \citep{huang2024mlagentbench}} & \small{Machine learning}     & N/A & - & - & - & No & No & 13    \\
    \small{SWE-Bench \citep{jimenez2024swebench}} & \small{Software engineering} & N/A & - & - & - & No & No & 2,294 \\
    \noalign{\hrule height 1pt}
    \multicolumn{9}{l}{\small{\textbf{Same Domain}}}\\
    \noalign{\hrule height 1pt}
    
    \small{DS-1000 \citep{lai2023ds}}              & \small{Data Science} & \small{Manually}      & \small{N/A}    & \small{N/A} & \small{No}  & \small{Yes} & \small{No}      & \small{1,000} \\
    \small{QRData \citep{liu-etal-2024-llms}}      & \small{Data Science} & \small{Manually}      & \small{15,186} & \small{46}  & \small{Yes}   & \small{No}  & \small{No}      & \small{411}   \\
    \small{Arcade \citep{yin2023natural}}          & \small{Data Science} & \small{Notebook Cells}& \small{N/A}    & \small{N/A} & \small{-}   & \small{No}  & \small{No}      & \small{1,078} \\
    \small{Spider2V \citep{cao2024spider2}}        & \small{Data Science} & \small{Manually}      & \small{N/A}    & \small{N/A} & \small{-}   & \small{-}   & \small{Yes}     & \small{494}   \\
    \small{DSEval \citep{zhang2024benchmarking}}   & \small{Data Science} & \small{Manually}      & \small{1,544}  & \small{12}  & \small{-}   & \small{No}  & \small{Partial} & \small{827}   \\
    \small{DSBench \citep{jingdsbench}}            & \small{Data Science} & \small{Data Files}    & \small{N/A}    & \small{N/A} & \small{-}   & \small{No}  & \small{No}      & \small{466}   \\
    \small{DA-Code \citep{huang2024code}}          & \small{Data Science} & \small{Manually}      & \small{9,639}  & \small{11}  & \small{No}   & \small{No}  & \small{No}      & \small{500}   \\
    \small{DataSciBench \citep{zhangdatascibench}} & \small{Data Science} & \small{Manually}      & \small{N/A}    & \small{N/A} & \small{-}   & \small{No}  & \small{No}      & \small{222}   \\
    \small{DABstep \citep{egg2025dabstep}}         & \small{Data Science} & \small{Manually}      & \small{N/A}    & \small{N/A} & \small{-}   & \small{No}  & \small{No}      & \small{450}   \\ 
    \noalign{\hrule height 1pt}
    \textbf{Ours (DSBC)} & \textbf{Data Science} & Structured & 7,793 & 10 & Yes & Yes & Yes & 303 \\    
    \noalign{\hrule height 1pt}
    \end{tabular}
    }
    \caption{Overview of some similar prior works in other domains and other existing Data Science benchmarks.}
    \label{table:1}
\end{table*}
\section{Related Works}
\label{sec:realtedworks}
\begin{figure*}[!h]
    \centering
    \includegraphics[width=1\linewidth]{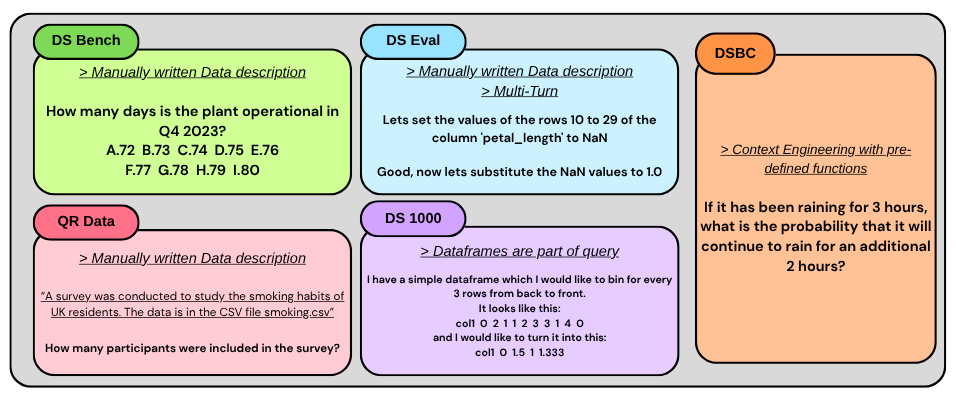}
    \caption{An example each from our benchmark along with some works cited above which were published less than 2 years ago.}
    \label{figure:09}
\end{figure*}

An overview of key differences between our benchmark and other benchmarks of the same domain can be seen in \autoref{table:1}. Examples of samples for each of the benchmarks can be seen in \autoref{figure:09}.

\paragraph{Other similar domain benchmarks:} 
Several prior works exist that benchmark code generation and other closely related tasks. One such prominent one is HumanEval \citep{chen2021evaluating} for code generation from text descriptions, along with Spider \citep{yu2018spider}, MBPP \citep{austin2021program}, and APPS \citep{hendrycks2measuring}. Other similar works include software engineering task benchmarks like automating code/PR review \citep{yang2016mining, li2022automating, tufano2023automating}, bug localization \citep{kim2019precise, chakraborty2018entropy}, testing \citep{kang2023large, xia2024fuzz4all, wang2024software}, program repair \citep{xia2022less, fan2023automated, sobania2023analysis}, and guiding code-editing \citep{chakraborty2021multi, zhang2022coditt5, fakhoury2023towards}.

\paragraph{Data-Science Benchmarks : } 
DS-1000 \citep{lai2023ds} is one of the earliest works for data science task evaluation with simple data analysis tasks sourced from StackOverflow; however, these questions lack usage of data files and require simpler 1-2 line code for many samples. DABstep \citep{egg2025dabstep} contains a fixed set of data files used for all samples of the benchmark with manually written data description. DataSciBench \citep{zhangdatascibench} consists of prompts that resemble text-to-code instructions, with the query itself providing context on necessary data file information. DA-Code \citep{huang2024code} also consists of queries that resemble text-to-code tasks, as the queries themselves provide step-by-step instructions on what steps to take to solve the given query. DS-Bench \citep{jingdsbench} contains tasks that are primarily designed for Excel-based workflows sourced from Modeloff competitions. DSEval \citep{zhang2024benchmarking} covers queries whose solutions mostly range from 1 to 3 lines of code, with a subset of questions being Leetcode problems. Arcade \citep{yin2023natural} consists of tasks in a notebook environment, while the previous cell of code provides the required context. QRData \citep{liu-etal-2024-llms} is a benchmark that closely resembles our benchmark with manually written data file descriptions appended to input prompts to provide the models with context. However, a majority of questions were MCQs. 
\section{Our Benchmark}
\label{sec:ourbenchmark}
Most prior benchmarks either have sourced or manually created descriptions of the data files used for the task, which are passed along with the query to the evaluation models. Few other works use data files of limited size that can be directly added in the context. This introduces randomness in the results based on how the data files' descriptions are written, what information is added, and what is withheld.

\begin{table}
    \centering
    \begin{tabular}{cc}
    \noalign{\hrule height 1pt}
    \rowcolor{orange} \textbf{\small{No.of Task types}} & \textbf{\small{Sample Count}} \\
    \noalign{\hrule height 1pt}     
    \small{One task categories}     & \small{63}  \\
    \small{Two task categories}     & \small{154}  \\
    \small{Three task categories}   & \small{86}  \\
    \noalign{\hrule height 1pt}     
    \end{tabular}
    \caption{Multi-label task category statistics of the queries (303) in our benchmark}
    \label{table:2}
\end{table}

The benchmark consists of 303 questions, which span 8 categories of tasks, with most questions covering more than one type of task. The categories and their descriptions, as well as their sample counts in the benchmark, can be seen below. \autoref{table:2} shows the number of task categories each sample in the benchmark spans.
\begin{itemize}
    \item \textbf{Correlation Analysis (44) : }Computing and interpreting relationships between variables using correlation coefficients, covariance matrices, and statistical significance testing.
    \item \textbf{Statistics (172) : }Performing descriptive and inferential statistical operations including confidence intervals, probability calculations, and computing mean, min, max, median etc.
    \item \textbf{Data Parsing \& Understanding (113) : }Interpreting data structure, content patterns, and contextual meaning to infer data origins, identify column semantics, and extract implicit information from datasets.
    \item \textbf{Data Pre-processing (69) : }Cleaning and preparing raw data through handling missing values, removing duplicates, removing empty rows and columns, removal of redundant features, outlier detection, and basic data quality assurance.
    \item \textbf{Feature Engineering (91) : }Creating new variables and features from existing data through mathematical transformations, aggregations, binning, cumulative and rolling features, and domain-specific feature construction.
    \item \textbf{Feature Transformation (85) : }Applying scaling, normalization, encoding categorical variables, dimensionality reduction, rounding and mathematical transformations.
    \item \textbf{Distribution Analysis (33) : }Examining data distributions through descriptive statistics, probability density functions, normality tests, and distributional property assessment.
    \item \textbf{Data Visualization (22) : }Creating charts, plots, and visual representations to explore patterns, communicate insights, and present analytical findings effectively.
\end{itemize}

\paragraph{Annotation : }
\begin{figure}
    \centering
    \includegraphics[width=1\linewidth]{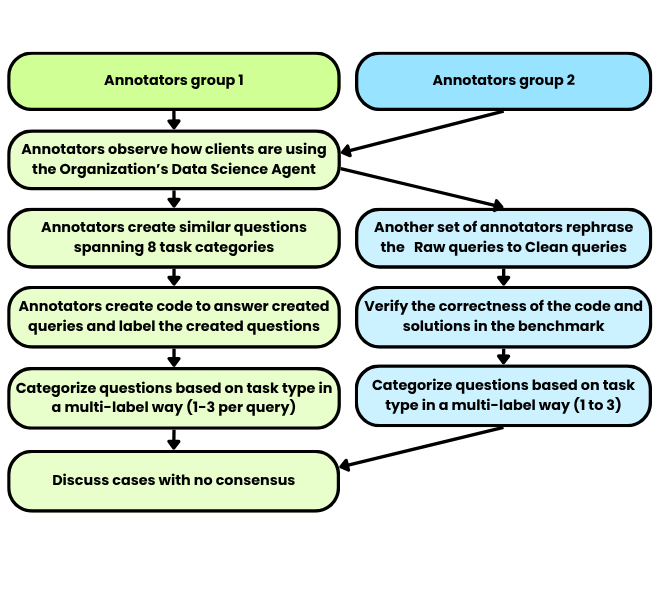}
    \caption{An overview of the annotation process used in our benchmark construction}
    \label{figure:10}
\end{figure}
The overview of the annotation process can be seen in \autoref{figure:10}. The annotators have created the dataset samples by observing how the clients were using the organization's data science agents. The distribution of task types was made to be close to the observed task type usage distribution. Queries were annotated, with their solutions manually coded and verified. Additionally, a copy of the raw queries was made to avoid data leakage, i.e., the clean queries. During annotation of task types for each query, the annotators were instructed to assign 1-3 task category tags to each query. In case the query appeared to be spanning more than 3 categories, they were instructed to assign the closest three. Samples where there was no consensus among the 2 sets of annotators were later manually verified after further discussion. The samples with no consensus were 26 out of 303, i.e., approximately 8.5\%. More details about annotator guidelines can be found in \autoref{sec:Annotationguidelines}.
\section{Context Engineering} 
\label{sec:contextengineering}
Organizations handling sensitive data face significant challenges when working with externally deployed or proprietary models. Direct file sharing raises privacy and security concerns, making it impractical to include raw data files in model contexts (even if they fit within the context limit). A structured approach that extracts only essential metadata, such as column counts, data types, categorical values, and temporal ranges, provides necessary context while maintaining data confidentiality and compliance requirements. 

Further, manual dataset documentation becomes increasingly burdensome as organizations scale their data operations. Writing detailed descriptions for hundreds or thousands of data files is time-consuming and resource-intensive. While automated description generation using LLMs offers a potential solution, it often produces incomplete or inaccurate characterizations that can mislead downstream analysis and compromise benchmark reliability.

We hence use context engineering to provide the models with the data files' description in a structured format covering several features like row and column count, column names, and data types, among other features. This can be seen in detail in \autoref{table:6}, which describes the features used in the context as a nested JSON dictionary.  

\begin{table*}
    \centering
    \begin{tabular}{cc}
    \noalign{\hrule height 1pt} 
    \rowcolor{orange} \textbf{Item}           &  \textbf{Description}\\
    \noalign{\hrule height 1pt}      
    Rows                    &  Number of Rows whether index is correctly ordered \\
    Columns                 &  Numbers of Columns and names of columns \\
    Data Types              &  Data types of each of the columns \\
    Null Counts             &  Null value counts of each of the columns \\
    Numeric Summary         &  If the column is Numeric then the Minimum, Maximum, Mean, \\
                            &  Median, 25th percentile value, 75th percentile value \\
    Categorical Summary     &  If the column is Categorical or has low unique \\
                            &  count (i.e <20), then the unique value counts \\
    DateTime Summary        &  If the column is already DateTime type, then \\ 
                            &  the start and end dates and whether the \\
                            &  column values' frequency is uniform \\  
    Sample rows **          &  First 5 rows' data of each data file \\                        
    \noalign{\hrule height 1pt}      
    \end{tabular}
    \caption{Features and description of info added through Context engineering}
    \caption*{** If the data file is of private nature, this is excluded}
    \label{table:6}
\end{table*}

\section{Datasets used}
\label{sec:datafiles}
The benchmarks utilize 11 different data files sourced from Kaggle, each of which covers a different domain: Farm produce data \citep{farm_produce_data}, Walmart sales \citep{walmart_sales}, COVID-19 mortality data \citep{covid19_dataset}, weather datasets \citep{weather_data}, insurance claims dataset \citep{insurance_data}, stock price datasets \citep{stock_market_dataset}, food inflation data \citep{monthly_food_price}, world population stats \citep{world_population_forecast}, air quality data \citep{air_quality_india}, electricity load data \citep{electricity_load_forecasting}, and life expectancy datasets \citep{life_expectancy_data}. The statistics about the data files can be seen in \autoref{table:5}.

\paragraph{Key differences : }
The data files chosen for the benchmarks have tricky features, which would make the benchmarks' samples more challenging compared to the rest. For instance, 8 of the 11 data files have a date or time column but not in a date-time datatype. The models/agents were provided with the first 5 rows of the data file through context engineering and are required to comprehend and figure out whether they need to perform a datatype conversion to successfully solve a query. One such tricky feature is the change of data frequency. The frequency of data for the population dataset changes midway from once every 5 years to yearly. Using the provided unique values list, the model/agent is required to figure out the necessary changes in the approach towards solving a query. Issues like these make the current benchmark more challenging than prior works. 

\begin{table*}[!h]
    \centering
    \resizebox{\textwidth}{!}{\
    \begin{tabular}{ccccccccccc}
    \noalign{\hrule height 1pt}
    \rowcolor{orange} \textbf{Dataset} & \textbf{Rows} & \textbf{Cols} & \textbf{Null.cols} & \textbf{DT} & \textbf{STR} & \textbf{CAT} & \textbf{BOOL} & \textbf{INT} & \textbf{FLT} & \textbf{AVG.ACC} \\
    \noalign{\hrule height 1pt}     
    Insurance   & 1,200   & 7  & 0  & 0 & 3 & 5  & 2 & 2  & 2  & 71.78 \\
    Weather     & 8,616   & 8  & 0  & 0 & 2 & 3  & 0 & 2  & 4  & 53.02 \\
    Power       & 3,624   & 17 & 0  & 0 & 1 & 3  & 2 & 3  & 13 & 48.06 \\
    COVID       & 10,000  & 21 & 0  & 0 & 1 & 19 & 3 & 20 & 0  & 45.31 \\
    AQI         & 8,737   & 23 & 21 & 0 & 2 & 3  & 0 & 0  & 21 & 41.60 \\
    Inflation   & 4,548   & 8  & 5  & 0 & 3 & 2  & 0 & 0  & 5  & 40.33 \\
    Sales       & 409,695 & 5  & 0  & 0 & 1 & 2  & 1 & 2  & 1  & 36.92 \\
    Health      & 13,942  & 5  & 0  & 0 & 2 & 1  & 0 & 1  & 1  & 34.78 \\
    Stocks      & 4,932   & 7  & 0  & 0 & 2 & 1  & 0 & 1  & 4  & 33.08 \\         
    Population  & 3,290   & 14 & 3  & 0 & 5 & 2  & 0 & 6  & 3  & 30.91 \\
    Production  & 9,464   & 9  & 6  & 0 & 1 & 3  & 0 & 2  & 5  & 27.34 \\
    \noalign{\hrule height 1pt}
    \end{tabular}
    }
    \caption{Features of datasets used for our benchmark : Row Count (ROW), Column Count (COLS), No.of columns with at least one null value (NULL.COLS), No.of Datetime (DT),String (STR), Categorical (CAT), Boolean (BOOL), Integer (INT) and float (FLT) columns respectively and the average score obtained from all attempts and temperature values over the queries using that dataset (AVG)}
    \label{table:5}
\end{table*}
\section{Query Types}
\label{sec:promptsandissues}
The raw (original) queries were rewritten (referred to as clean queries) to ensure no data leakage occurs even though the impact is negligible. Though the effect could be negligible, this was done to compare the change in results with both sets of queries. Most queries of usage of our commercial agents resembled the format of raw queries. \autoref{table:3} shows an example for the raw and clean query of one of the samples. Another difference between Raw and Clean queries is that Clean prompts require an SQL-like approach, as demonstrated in \autoref{table:4}.

\begin{table}
    \centering
    \begin{tabular}{cc}
    \noalign{\hrule height 1pt} 
    \rowcolor{orange} \multicolumn{2}{c}{\textbf{Queries Example 1}}\\
    \noalign{\hrule height 1pt} 
    \textbf{Raw}     & Among those who have died, \\
                     & how many deaths were not  \\
                     & attributed to COVID-19 ?  \\
    \noalign{\hrule height 1pt}      
    \textbf{Clean}   & Did any deaths occur not due to \\
                     & COVID-19? If so, how many? \\
    \noalign{\hrule height 1pt}      
    \end{tabular}
    \caption{An example of Raw and Clean prompts : The raw prompt assumes deaths occurred (data leakage), while the clean prompt requires the model to check if deaths happened first.}
    \label{table:3}
\end{table}

\begin{table}
    \centering
    \begin{tabular}{cc}
    \noalign{\hrule height 1pt} 
    \rowcolor{orange} \multicolumn{2}{c}{\textbf{Queries Example 2}}\\
    \noalign{\hrule height 1pt} 
    \textbf{Raw}     & Which three countries have had \\
                     & the most stable fertility rates ?\\
    \noalign{\hrule height 1pt}      
    \textbf{Clean}   & Which countries have had the \\
                     & most stable fertility rates?  \\
                     & List the top 3. \\
    \noalign{\hrule height 1pt}      
    \end{tabular}
    \caption{An example of Raw and Clean prompts : The raw prompt directly requests the top 3 countries, while the clean prompt follows a SQL-like approach of first identifying all stable countries, then selecting the top 3.}
    \label{table:4}
\end{table}
\begin{figure*}[!h]
    \centering
    \includegraphics[width=1\linewidth]{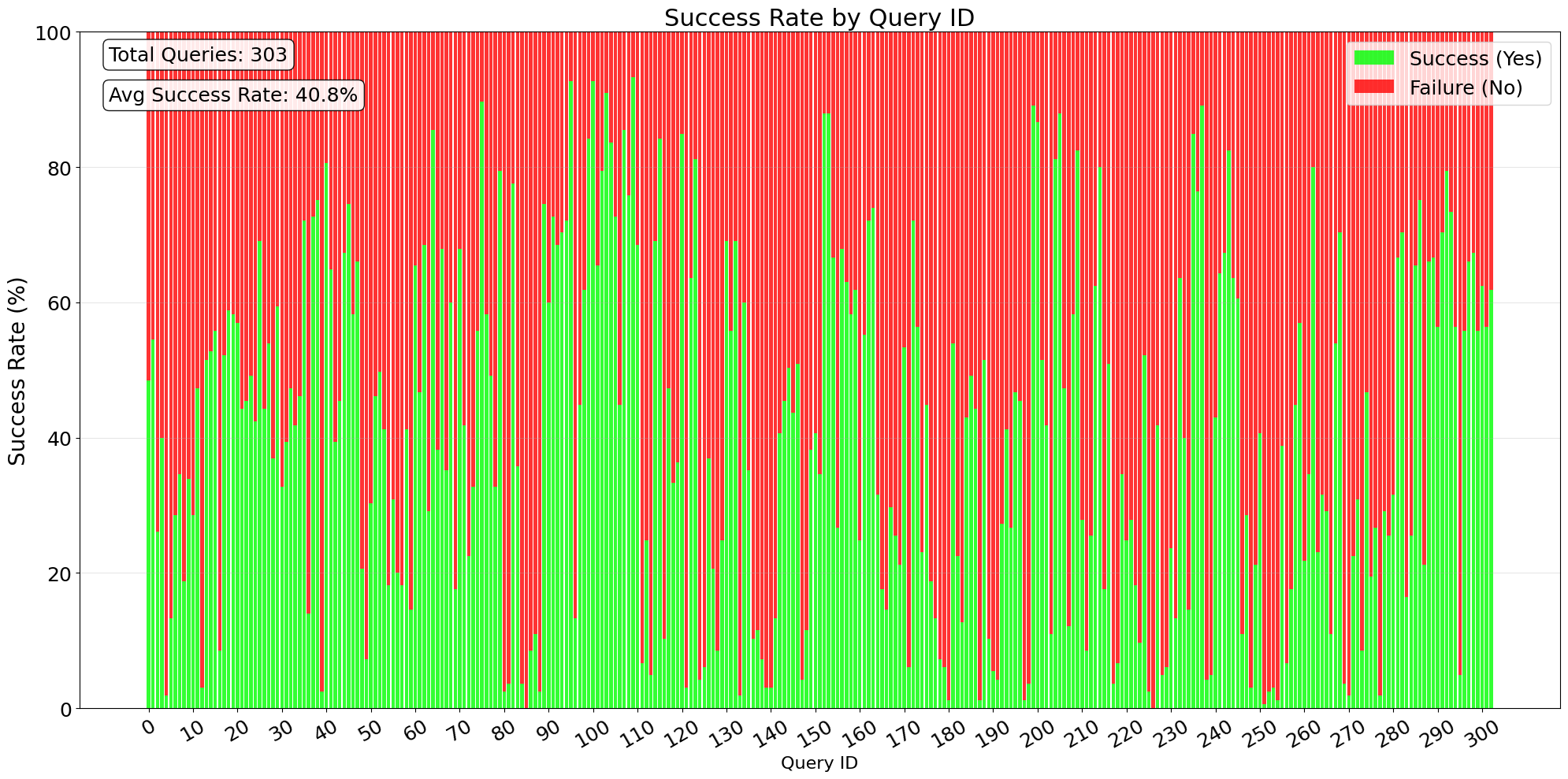}
    \caption{Success rates for each sample (303) of our benchmark over several (165) attempts}
    \label{figure:1}
\end{figure*}
\begin{figure*}[!h]
    \centering
    \includegraphics[width=0.95\linewidth]{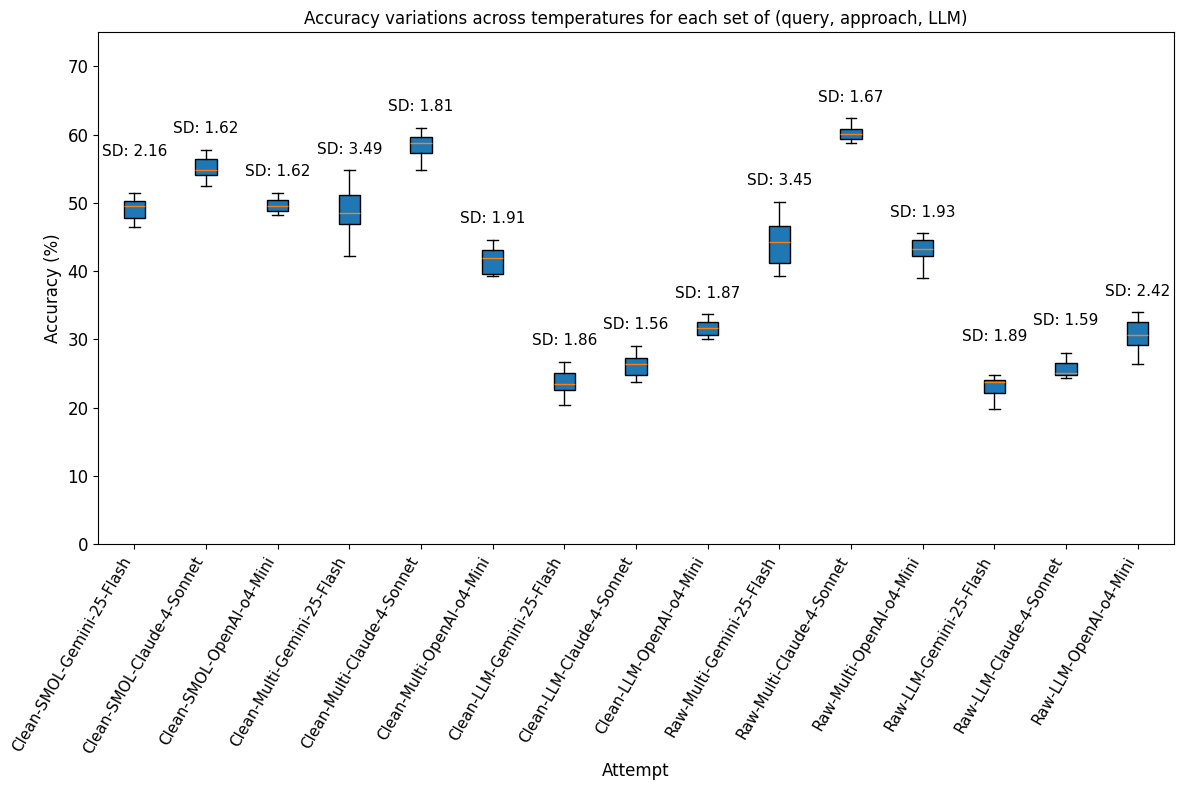}
    \caption{Variation in Accuracies through each set of (LLM, query type and approach)}
    \label{figure:4}
\end{figure*}
\section{Evaluation}
\label{sec:baselineevals}
We evaluate the samples using 3 models (Claude-4-Sonnet, Gemini-2.5-Flash, and OpenAI-o4-Mini) in 3 settings (directly with context engineering with LLMs (ReACt), multi-step with context engineering using LLMs (ReAct), and SmolAgent) and through 2 query types (raw and clean). These 15 unique setups \footnote{SmolAgent was initially tested with 1 randomly chosen temperature value over all 3 models; no change was observed in results between the Raw and Clean queries. Hence, due to a limited compute budget, SmolAgent was only tested on clean queries for all temperature values.} were tested over 11 temperature values ranging from 0.0 to 1.0 in 0.1 increments. This resulted in a total of 165 attempts over each of the samples whose results are described below.

\begin{table*}[!h]
    \centering
    \resizebox{\textwidth}{!}{\
    \begin{tabular}{ccccccccc}
    \noalign{\hrule height 1pt} 
    \rowcolor{orange} \textbf{LLM used} & \textbf{Approach Used} & \textbf{Query Type} & \textbf{Highest}  & \textbf{Highest at}  & \textbf{Lowest}   & \textbf{Highest at}  & \textbf{Overall}  & \textbf{Standard}  \\
    \rowcolor{orange} \textbf{}         & \textbf{}              & \textbf{}           & \textbf{Accuracy} & \textbf{Temperature} & \textbf{Accuracy} & \textbf{Temperature} & \textbf{Accuracy} & \textbf{Deviation} \\
    \noalign{\hrule height 1pt}      
    Claude-4-Sonnet     & SmolAgent   & Clean & 57.756 & 0.0 & 52.475 & 0.5 & 55.236 & 1.625 \\
    OpenAI-o4-Mini      & SmolAgent   & Clean & 51.485 & 0.7 & 45.215 & 1.0 & 49.385 & 1.618 \\
    Gemini-2.5-Flash    & SmolAgent   & Clean & 54.455 & 0.0 & 46.535 & 0.1 & 49.415 & 2.162 \\
    \noalign{\hrule height 1pt}      
    Claude-4-Sonnet     & Multi-Code  & Clean & 61.056 & 0.9 & 54.785 & 0.1 & 58.356 & 1.812 \\
    Claude-4-Sonnet     & Multi-Code  & Raw   & 62.376 & 0.2 & 55.776 & 0.5 & 59.916 & 1.673 \\
    OpenAI-o4-Mini      & Multi-Code  & Clean & 44.554 & 0.4 & 39.274 & 0.8 & 41.524 & 1.913 \\
    OpenAI-o4-Mini      & Multi-Code  & Raw   & 45.545 & 0.9 & 38.944 & 0.7 & 43.054 & 1.931 \\
    Gemini-2.5-Flash    & Multi-Code  & Clean & 54.785 & 0.9 & 42.244 & 0.7 & 49.055 & 3.488 \\
    Gemini-2.5-Flash    & Multi-Code  & Raw   & 50.165 & 0.7 & 39.274 & 0.5 & 44.164 & 3.452 \\
    \noalign{\hrule height 1pt}      
    Claude-4-Sonnet     & Single-Code & Clean & 29.043 & 0.3 & 23.762 & 0.4 & 26.163 & 1.561 \\
    Claude-4-Sonnet     & Single-Code & Raw   & 29.703 & 0.3 & 24.422 & 0.8 & 25.953 & 1.594 \\
    OpenAI-o4-Mini      & Single-Code & Clean & 33.663 & 0.1 & 26.733 & 0.4 & 31.443 & 1.873 \\
    OpenAI-o4-Mini      & Single-Code & Raw   & 33.993 & 0.1 & 26.403 & 0.2 & 30.633 & 2.420 \\
    Gemini-2.5-Flash    & Single-Code & Clean & 26.733 & 0.0 & 20.462 & 1.0 & 23.792 & 1.859 \\
    Gemini-2.5-Flash    & Single-Code & Raw   & 27.393 & 0.7 & 19.802 & 0.5 & 23.342 & 1.889 \\
    \noalign{\hrule height 1pt}      
    \end{tabular}
    }
    \caption{Results from each of the 15 attempts : Overall Accuracy and Standard Deviation with varying temperature}
    \label{table:7}
\end{table*}

The generated output and explanation/rationale by the model/agent were evaluated using VLM-as-Judge using Gemini-2.5-Flash. Three of the 165 attempts (one from each LLM) were checked manually to see whether the VLM-as-a-judge is indeed evaluating the responses accurately. JSON schema was used for the VLM-as-a-judge setup, where the VLM responds with a single word, either 'Yes' or 'No,' based on whether the response is accurate. To account for noise, these responses were parsed through regex as a final step.

The success rate (the percentage of attempts (165) that resulted in an accurate response) had a mean of 40.8\% and a median of 41.21\%. No question was answered correctly in all attempts. Only two questions were answered incorrectly in all attempts, both of which are reasonably hard questions. These samples and why the models/agents always resulted in an incorrect response were elaborated in detail in \autoref{sec:unsolvedqueries}. The maximum success rate was 93.33\%, and only 23 samples had a success rate > 80\%. More can be seen in \autoref{figure:1}.
\subsection{Using LLMs Directly with Context Engineering}
\label{subsec:evalsapproach1}
In this approach, the execution was done in a single-step ReAct \citep{yao2023react} loop with sandboxed code execution, where the LLM generates code that takes the query and added context as input; the code is then executed locally, which becomes the final output. When generating the code, the LLM is also instructed to add its rationale for the generated code. The rationale and code were returned as a JSON, which are then individually parsed through predefined functions. 
\subsection{Using LLMs in multiple steps with Context Engineering}
\label{subsec:evalsapproach2}
This approach is similar to the previous approach, but with one minor change: the code and explanations generated were divided into 2-3 steps. Unlike the traditional ReAct workflow, in this approach the LLM generates 2-3 code snippets (one for each step) to solve the query in one go, which are then executed one by one. The rationale is generated separately for each snippet, which are then appended to one another in the end.
\subsection{Using SmolAgent}
\label{subsec:evalsapproach3}
In this approach, we use SmolAgent's CoderAgent out of the box with minor changes: adding the data files to the execution environment's runtime and adding the path to the files in the query. No additional context is provided to the agent. The agent then uses some of its multiple steps allowed to iteratively gain context on what the data files contain and answers the query in subsequent attempts. A computational and time limit was assigned at 8 steps and 90 seconds, respectively, for each sample. None of the samples' 33 attempts (3 LLMs * 11 temperature values) resulted in a timeout error.

\section{Results}
\label{sec:results}

\paragraph{Effect of temperature on results : } 
Results from each of the 15 attempts can be seen in \autoref{table:7}, which combines all samples across each temperature value. No significant pattern was seen among the change in accuracies versus each temperature value for all sets of prompt, model, and approach, as seen in \autoref{figure:2}. Claude-4-Sonnet, through multiple code snippets in a single step, clearly outperformed the rest, including SmolAgent with several steps, irrespective of temperature. Multi-step and SmolAgent performed significantly better than single-cell direct use of LLMs when looking at the overall results. However, when looking at the same results for each task category separately, there is a large variation in the same plot. This can be seen in \autoref{sec:taskwiseaccuracies}. 

\paragraph{Variations in results with Model and approach : } 
Gemini-2.5-Flash demonstrated greater sensitivity to temperature compared to the other two LLMs tested, especially with the multi-code-cell approach and through SmolAgent. For single code cell implementation, o4-mini clearly outperformed the other two LLMs, but the performance of o4-mini was closer to SmolAgent and lower than Gemini-2.5-Flash in multi-code-cell implementation, as seen in \autoref{figure:4}.

\paragraph{Variations in results with Model and approach : } 
Certain domains' samples had higher accuracy than the rest irrespective of model, approach, or query type despite the code complexity and task difficulty being the same. The temperature values that produced the better results in each of the domains also varied by a considerable extent. 
\section{Error Analysis}
\label{sec:erroranalysis}
\begin{figure}[!h]
    \centering
    \includegraphics[width=1\linewidth]{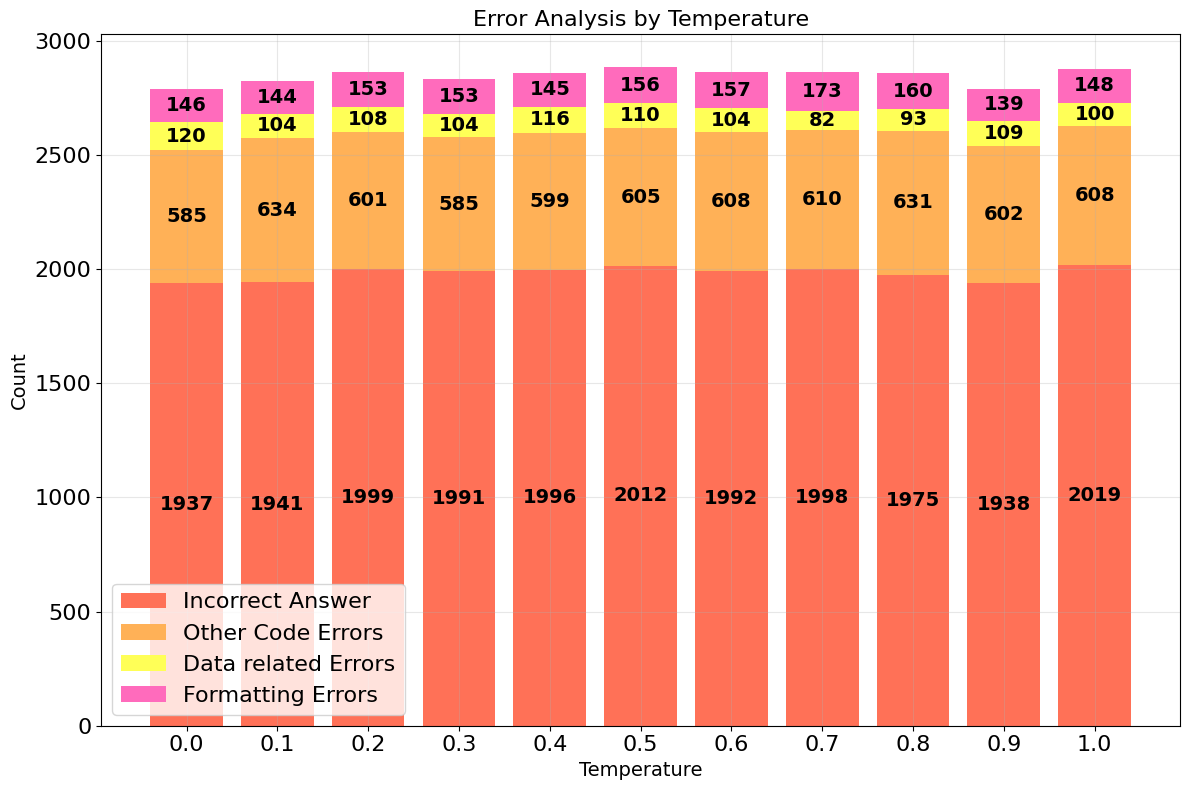}
    \caption{Error cause distribution overall : temperature wise}
    \label{figure:7}
\end{figure}

Among the unsuccessful cases, roughly 70\% were due to incorrect final responses, another 20\% were due to other errors in code, 6\% were due to data error from not being able to access the data, and the rest were due to formatting errors by the LLM in returning the code. This can be seen in \autoref{figure:7} for all attempts combined for each temperature value used. The formatting errors did not occur through SmolAgent due to the use of ReAct and are from the rest of the approaches. Formatting errors disproportionately occurred while using o4-mini as the LLM as compared to the other two LLMs. In the single-code-cell approach, the count of code errors outnumbered the count of instances where an incorrect response was generated. More on this can be seen in \autoref{sec:errorcounts}.

\begin{figure}
    \centering
    \includegraphics[width=1\linewidth]{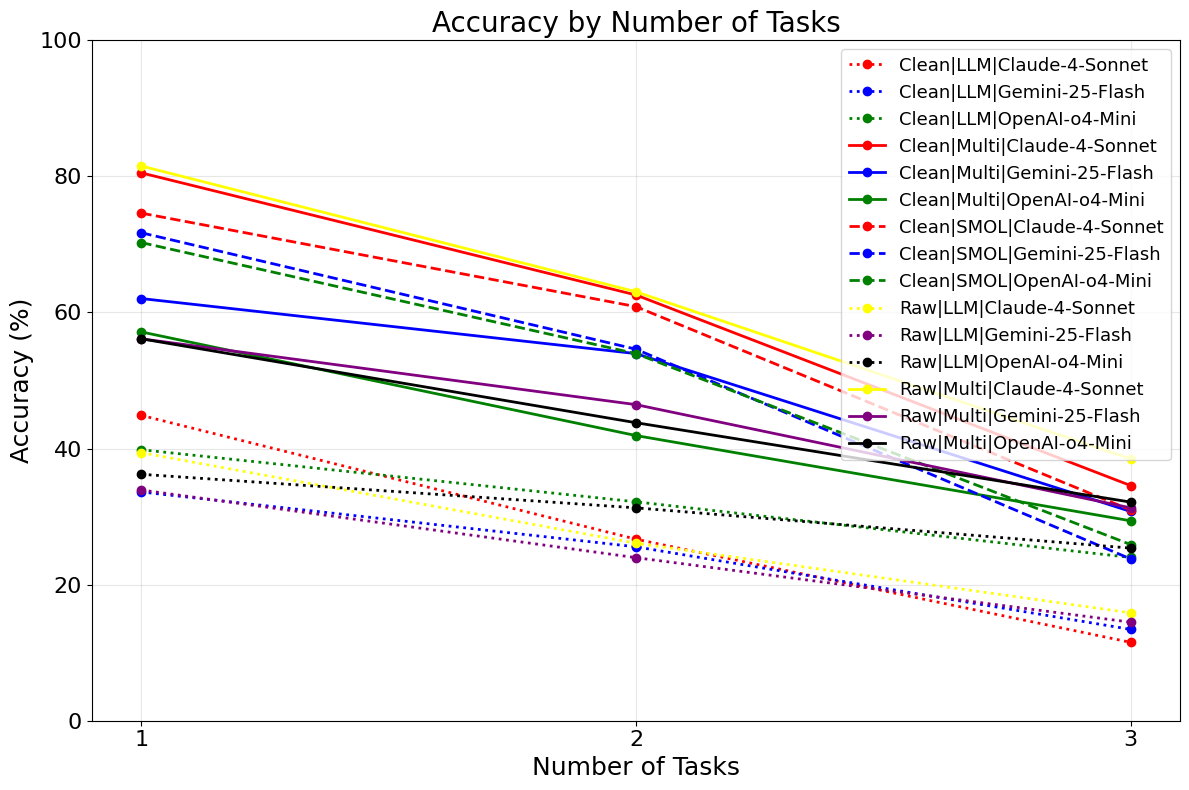}
    \caption{Average accuracy with each attempt VS number of task categories the query covers}
    \label{figure:11}
\end{figure}

\paragraph{Error rate on single and multi-task samples :}
The overall accuracy, including all attempts with all temperature values over samples that cover 1, 1,2 and 3 task categories, is 55.86\%, 43.13\%, and 25.46\%, respectively; i.e., the accuracy drops significantly if the query requires code that spans more than one task type, as in \autoref{sec:ourbenchmark}. The comparison of accuracies for each attempt over samples that span varying numbers of task types can be seen below in \autoref{figure:11}. Reasoning models like OpenAI-o4-mini have observed smaller drops in performance with increasing complexity of queries, i.e., queries that span several task types, while a non-reasoning model might be an efficient choice for simpler queries. Additionally, we have also seen that certain tasks benefit from expensive approaches (SmolAgent or multi-cell code), while many other tasks do not, as seen in \autoref{sec:taskwiseaccuracies}. This hints that an efficient query-based model routing could be built that can reduce costs incurred while retaining similar performance by solving simpler queries with a less expensive approach and LLM.

\section{Conclusion}
\label{sec:conclusion}
Through this paper, we introduce DSBC, a data science agent benchmark with context engineering and additional parallel prompts. The benchmark closely resembles the type and way of usage of the organization' commercial data science agent. The benchmark is being released through the Apache 2.0 license to facilitate further research in the domain of data science agents.
\section*{Limitations}
The benchmark does not include multilingual or multi-modal questions and is limited to text based queries. Multilingual queries can work through agents with additional steps for translation and back translation, however they haven't been tested.
\bibliography{anthology,custom}
\bibliographystyle{acl_natbib}
\appendix
\onecolumn
\section{Prompts Used}
\label{sec:promptsused}

\begin{tcolorbox}[width=\textwidth,colback={white},title={Prompt 1 : SmolAgent},colbacktitle=orange,coltitle=black]    
    \begin{verbatim}
INSTRUCTIONS : the final answer must contain the answer mentioned explicitly whether 
if it is a text, list of answers, something numeric or text in the desired format.

QUERY : 
{question}.

DATASET PATH : 
{dataset_path}
    \end{verbatim}
\end{tcolorbox}

\begin{tcolorbox}[width=\textwidth,colback={white},title={Prompt 2 : Multi-Code-Cell},colbacktitle=orange,coltitle=black]    
    \begin{verbatim}
You are a data analyst. The user asked: "{user_query}"
Dataset information:
- First 5 rows: {sample_rows}
- Dataset description: {dataset_description}

1. Generate COMPLETE, STANDALONE code snippets (2-3 steps) that 
could each run independently. For each step, provide:
1. A text explanation of what this step does
2. The Python code snippet that implements this step

Format your response as JSON with this structure:
{{
  "steps": [
    {{
      "explanation": "Detailed explanation of what this step accomplishes",
      "code": "Complete Python code with imports, analysis, and output"
    }}
  ]
}}

REQUIREMENTS:
- Use pandas to load and analyze the data: pd.read_csv('{filepath}')
- MUST include at least one visualization using matplotlib/seaborn
- Include necessary imports in each relevant snippet
- Complete analysis logic
- Clear visualization code when applicable
- Proper print() statements to show results
- Don't use markdown formatting, just pure JSON
    \end{verbatim}
\end{tcolorbox} 

\begin{tcolorbox}[width=\textwidth,colback={white},title={Prompt 3 : Explanation prompt},colbacktitle=orange,coltitle=black]    
    \begin{verbatim}
A user submitted the following question about their dataset: "{user_question}"

Dataset Context:
{dataset_info}
Generated Analysis Code:
{code}
Execution Results:
{answer}

Please provide a clear, comprehensive explanation that:
1. Directly addresses the user's original question
2. Interprets the results in the context of the dataset
3. Explains what the findings mean in practical terms
4. Highlights any key insights or patterns discovered
5. Always use first-person when speaking.

Your explanation should be written in plain language that a business 
stakeholder could easily understand.
    \end{verbatim}
\end{tcolorbox} 

\begin{tcolorbox}[width=\textwidth,colback={white},title={Prompt 4 : Single-Code-Cell},colbacktitle=orange,coltitle=black]    
    \begin{verbatim}
You are a data analyst. Generate Python code to analyze a CSV file and answer 
the following query: {user_query}

      Dataset information:
      {dataset_description}
      # # # # # # # # # # # # # # # # # # # # #
      Requirements:
      - Use pandas to load and analyze the data: pd.read_csv('data.csv')
      - Include necessary imports
      - Use print() statements to show results
      - For visualizations, use matplotlib/seaborn
      - Handle data type conversions if needed
      - Return ONLY executable Python code, no markdown formatting
      - The dataframe is already loaded in a variable called 'data'. Do not re-read it
      - Have the answer ready in a variable called 'answer'. 
      Just declare and add your values there.
      - Do not have subplots, only main plots
      Generate clean, executable Python code
      For the Explination, it should describe why a step was taken and not whats done.
      Use this format for the response :

      {{
          "explanation": "...",
          "code": "..."
      }}
    \end{verbatim}
\end{tcolorbox} 

\begin{tcolorbox}[width=\textwidth,colback={white},title={Prompt 5 : VLM-as-a-Judge},colbacktitle=orange,coltitle=black]    
    \begin{verbatim}
Respond in this exact JSON format:
{
  {
    "Evaluation": 'Yes' 'No' 'The Evaluation should be Yes only when the response is 
    technically correct. Sometimes the answer might be of a different format but 
    still correct (Ex : March , 3 when asked the month etc..). For numerical values, 
    the rounded .2f values should match to be considered correct.'
  }
}

You are being used for LLM-as-judge. In numeric solutions an error beyond the 2nd 
decimal point after rounding can be ignored.

####
The Query by the user is :
{Q}
----
The Ground Truth for the query is :
{A}
----
The Code Snippet to obtain the ground truth was :
{C}
----
The Response by the model is :
{R}
----
The Code Snippet in the submission was :
{S}
----
The Reasoning given with the submission was :
{N}
    \end{verbatim}
\end{tcolorbox} 
\section{Reproducibility}
\label{sec:Reproducibility}
\noindent
The hyperparameters not specified or tested with multiple values through the paper are all use through their default values. A max token limit of 8192 was used. All experiments were done on Google Cloud. All inference runs combined have cost approx. 1200\$, 200\$ and 700\$ respectively for Claude-4-Sonnet, Gemini-2.5-Flash and OpenAI-o4-Mini respectively. For VLM-as-a-judge experiments, it cost us around 150\$ using Gemini-2.5-Flash. For SmolAgent, we used a time limit of 90 seconds and step limit of 5 for each query during inference. Additional allowed imports were constrained to a few packages that should be enough to solve the queries of the benchmarks : pandas, numpy, scipy, scikitlearn, matplotlib, seaborn, re, math, datetime. For the same amount of inference samples, compared to single-code-cell, multi-code-cell was 1.8x expensive and SmolAgent was 3.2x expensive when averaging costs across all LLMs tested. All LLMs were used with a seed value of 1024.\\
\newpage
\section{Annotation Guidelines}
\label{sec:Annotationguidelines}
\begin{tcolorbox}[width=\textwidth,colback={white},title={Annotator Guidelines : Creating Samples},colbacktitle=orange,coltitle=black]    
    \begin{verbatim}
When creating training samples, focus on generating simple, realistic queries that 
reflect actual user interactions you've encountered in client work or observed in 
system usage logs. mirror the natural language patterns, terminology, and problem 
types that real users typically present. Draw from common scenarios you've 
witnessed, such as troubleshooting requests, feature questions, or workflow 
clarifications, ensuring that each sample captures the authentic tone and 
complexity level of genuine user inquiries rather than overly polished or 
artificial examples.
    \end{verbatim}
\end{tcolorbox} 

\begin{tcolorbox}[width=\textwidth,colback={white},title={Annotator Guidelines : Making Clean queries},colbacktitle=orange,coltitle=black]    
    \begin{verbatim}
When creating clean queries, remove language that leaks information or makes 
assumptions about the data. The query "Among those who have died, how many deaths 
were not attributed to COVID-19?" assumes non-COVID deaths exist in the dataset. 
A cleaner version asks "Did any deaths not occur due to COVID-19? If so, how many?" 
This removes the data leakage while keeping the same core question. Make sure that 
the exact meaning of both the Clean query and Raw query is the same.
    \end{verbatim}
\end{tcolorbox} 

\begin{tcolorbox}[width=\textwidth,colback={white},title={Annotator Guidelines : categorizing Samples},colbacktitle=orange,coltitle=black]    
    \begin{verbatim}
Categorize the queries based on what tasks they require to be performed to be 
able to answer the queries. A query can be categorizied as atleast one category 
and at most 3 categories. If the sample seems close to more than three categories, 
assign the best matching three categores, use the below examples for reference : 
{examples}.
    \end{verbatim}
\end{tcolorbox}

\newpage

\section{Unsolved Samples}
\label{sec:unsolvedqueries}
The two unsolved samples, its solution and the common errors made by the LLMs/agents were :

\begin{tcolorbox}[width=\textwidth,colback={white},title={Example 1 : Query ID 8},colbacktitle=orange,coltitle=black]    
   \textbf{If it rains today, what is the likelihood that it will rain tomorrow as well?}
    \begin{verbatim}
import pandas as pd
# Convert to datetime
df_AQI['From Date'] = pd.to_datetime(df_AQI['From Date'])
df_AQI['Date'] = df_AQI['From Date'].dt.date
# Daily rainfall sums
daily_rain = df_AQI.groupby('Date')['RF (mm)'].sum().reset_index()
daily_rain = daily_rain.sort_values('Date').reset_index(drop=True)
# Binary rain indicators (>0.1mm = rain)
daily_rain['Rain_Today'] = (daily_rain['RF (mm)'] > 0)
daily_rain['Rain_Tomorrow'] = daily_rain['Rain_Today'].shift(-1)
# Remove last row (no tomorrow data)
daily_rain = daily_rain[:-1]
# Calculate probability
rain_today_count = daily_rain['Rain_Today'].sum()
rain_both_days = ((daily_rain['Rain_Today'] == 1) 
                 & (daily_rain['Rain_Tomorrow'] == 1)).sum()
probability = rain_both_days / rain_today_count if rain_today_count > 0 else 0
print(f"P(Tomorrow | Today) = {probability:.3f} ({probability*100:.1f}%)")
\end{verbatim}
\textbf{Mistakes made by the LLMs and Agents include, A) blind trust in "From Date" column to be a date-time column without verifying, B) assuming "From Date" to cover daily data rather than hourly data without verifying, C) using rainfall >0.1 mm as rain rather than directly using 0.0, incorrectly assuming all rows are of .1f type}
\end{tcolorbox} 

\begin{tcolorbox}[width=\textwidth,colback={white},title={Example 2 : Query ID 226},colbacktitle=orange,coltitle=black]    
   \textbf{Which of the calendar months typically experience the highest sales in an year ? List top 2}
    \begin{verbatim}
df_SALES['Date'] = pd.to_datetime(df_SALES['Date'])
df_SALES['Month'] = df_SALES['Date'].dt.month
monthly_sales = df_SALES.groupby('Month')['Weekly_Sales'].mean()
sorted_monthly_sales = monthly_sales.sort_values(ascending=False)
top_2_months = sorted_monthly_sales.head(2)
month_names = { 1: 'January', 2: 'February', 3: 'March', 4: 'April',
                5: 'May', 6: 'June', 7: 'July', 8: 'August',
                9: 'September', 10: 'October', 11: 'November', 12: 'December'}
top_2_month_names = [month_names[month] for month in top_2_months.index]
print(f"{top_2_month_names[0]} and {top_2_month_names[1]}")
\end{verbatim}
\textbf{Mistakes made by the LLMs and Agents include, A) grouping 2 month periods from start rather than considering all possible rolling 2 month periods B) assuming dataset starts and ends in same month. i.e if it starts in May for example and ends in July of a different year. There is more data for June and hence considering sum() instead of mean() leads to a incorrect response. C) assuming the question meant which 2 months despite being asked for "calendar months"}
\end{tcolorbox} 
\newpage
\section{Task wise Success Rates}
\label{sec:successrates}
The task wise success rates of each query spanning 165 attempts used can be seen in \autoref{figure:03A}, \autoref{figure:03B}, \autoref{figure:03C}, \autoref{figure:03D}, \autoref{figure:03E}, \autoref{figure:03F}, \autoref{figure:03G}, \autoref{figure:03H} respectively for each task category separately. The overall success rates for the several task categories ranged from a lowest value of of 29.1\% for Feature engineering to a highest value of 46.9\% for Correlation analysis. the distributions of each of the tasks' success rates are considerably skewed half each of either direction with differences between mean and median ranging from 4-15\% among the samples of the same task category.
\begin{figure*}[!h]
    \centering
    \includegraphics[width=1\linewidth]{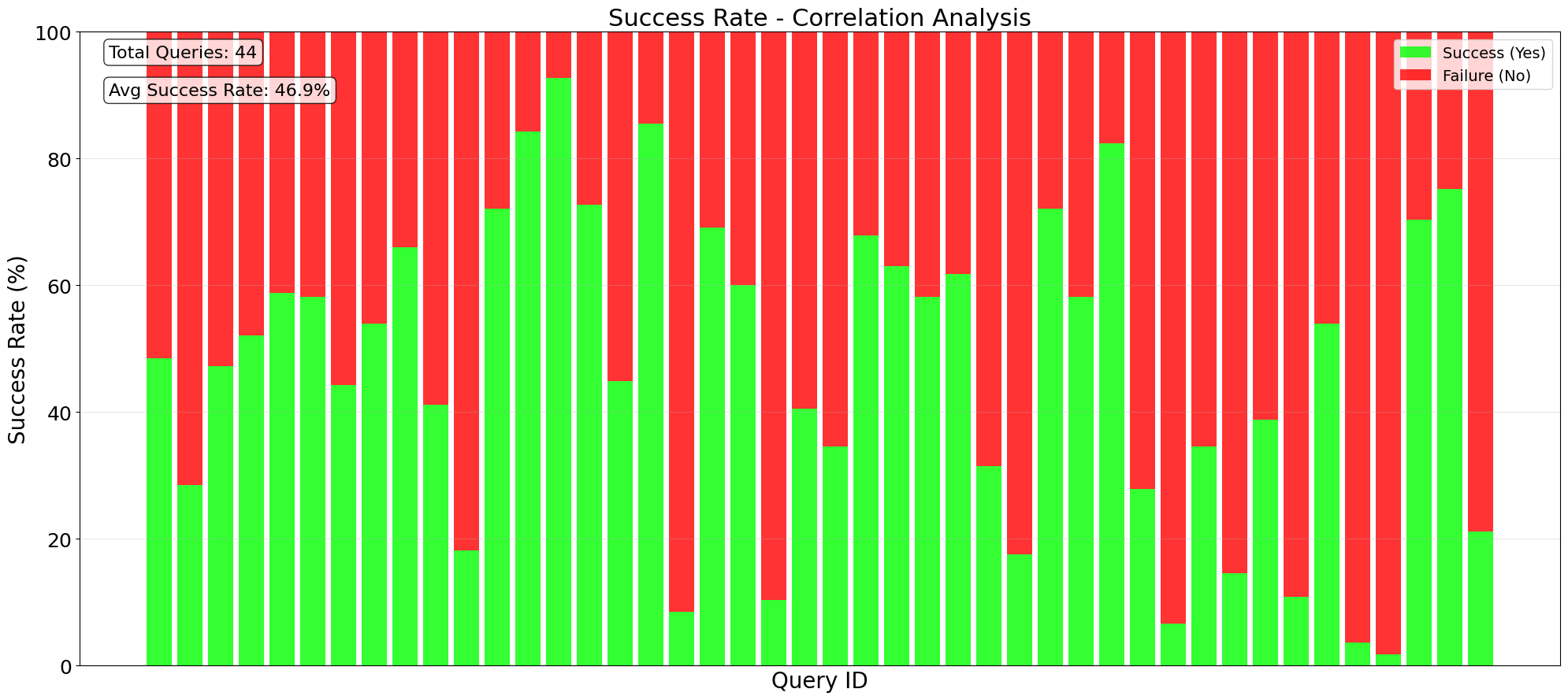}
    \caption{Success rates for each sample over several (165) attempts - Correlation Analysis}
    \label{figure:06A}
\end{figure*}
\begin{figure*}[!h]
    \centering
    \includegraphics[width=1\linewidth]{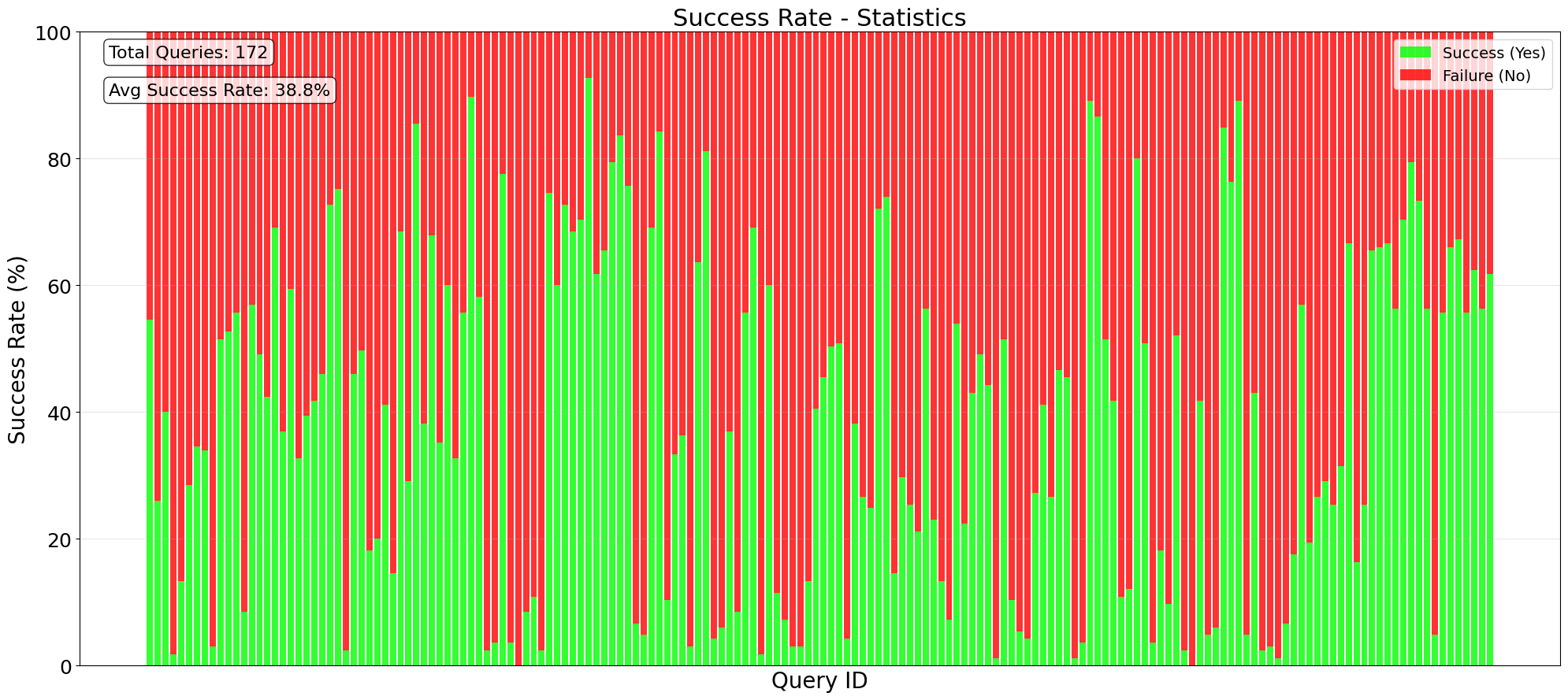}
    \caption{Success rates for each sample over several (165) attempts - Statistics}
    \label{figure:06B}
\end{figure*}
\newpage
\begin{figure*}[!h]
    \centering
    \includegraphics[width=0.93\linewidth]{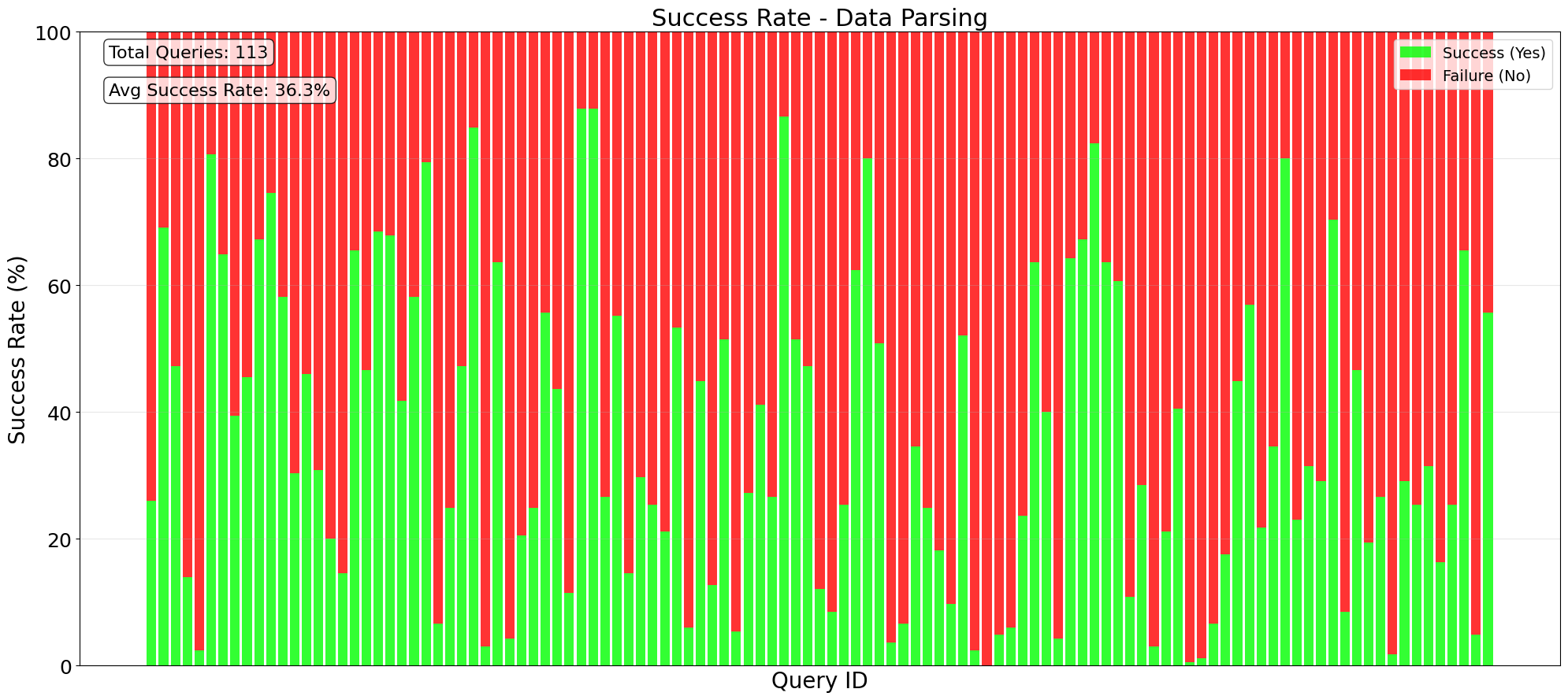}
    \caption{Success rates for each sample over several (165) attempts - Data Parsing}
    \label{figure:06C}
\end{figure*}
\begin{figure*}[!h]
    \centering
    \includegraphics[width=0.93\linewidth]{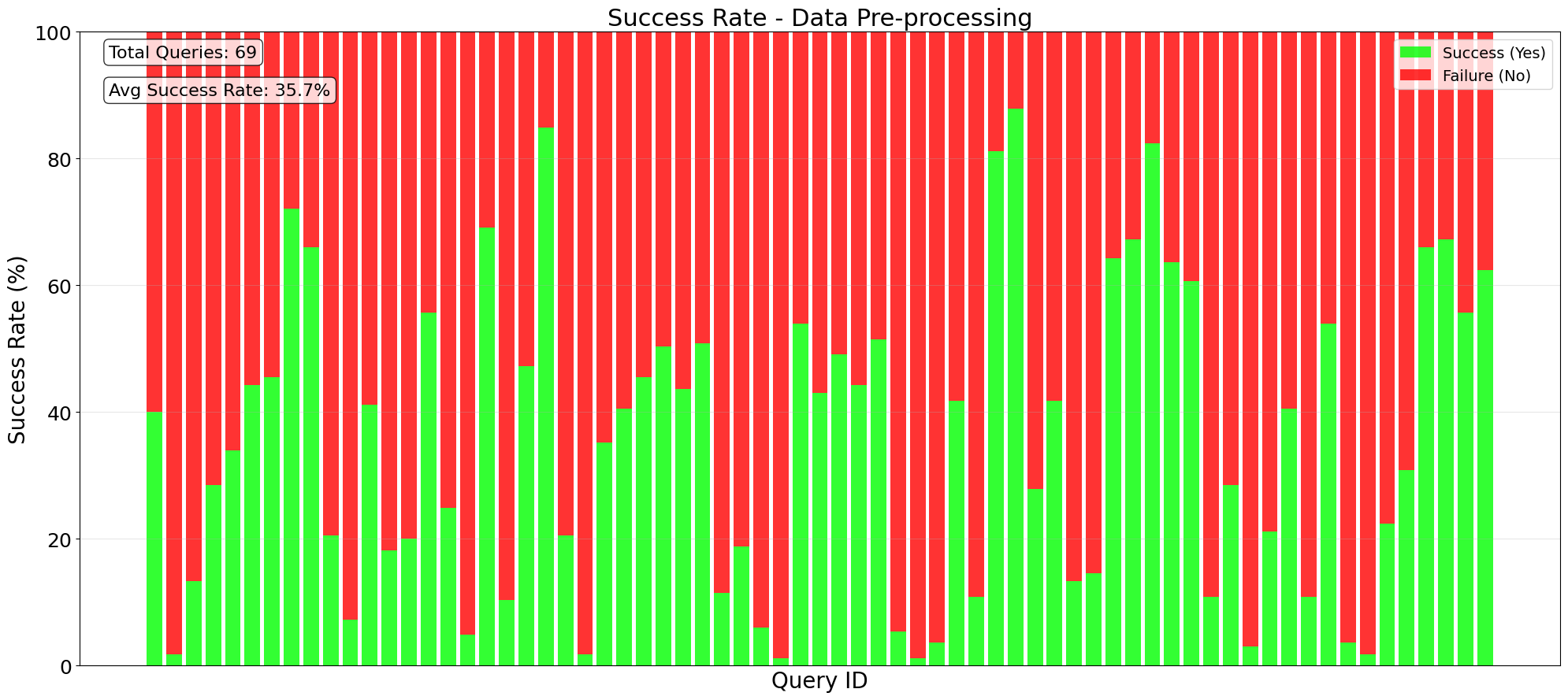}
    \caption{Success rates for each sample over several (165) attempts - Data Pre-processing}
    \label{figure:06D}
\end{figure*}
\begin{figure*}[!h]
    \centering
    \includegraphics[width=0.93\linewidth]{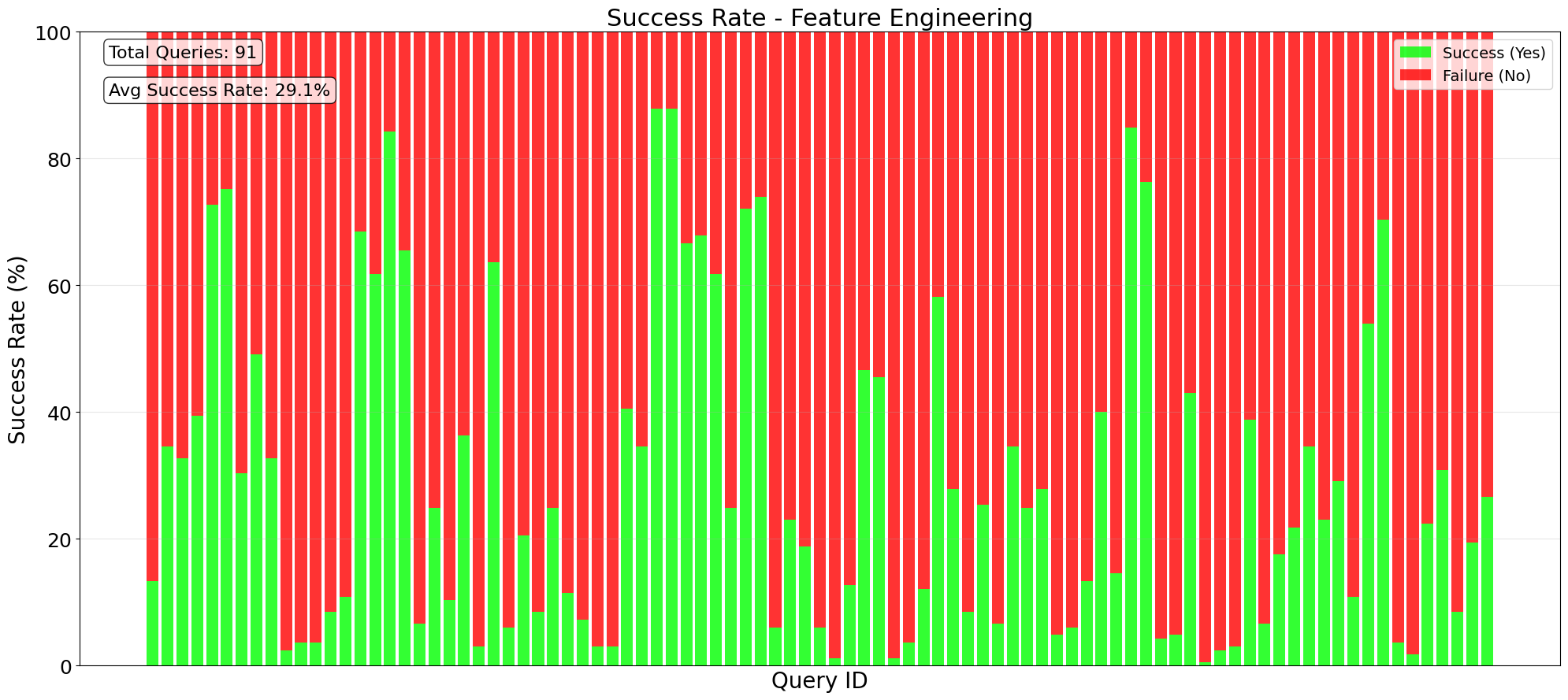}
    \caption{Success rates for each sample over several (165) attempts - Feature Engineering}
    \label{figure:06E}
\end{figure*}
\newpage
\begin{figure*}[!h]
    \centering
    \includegraphics[width=0.93\linewidth]{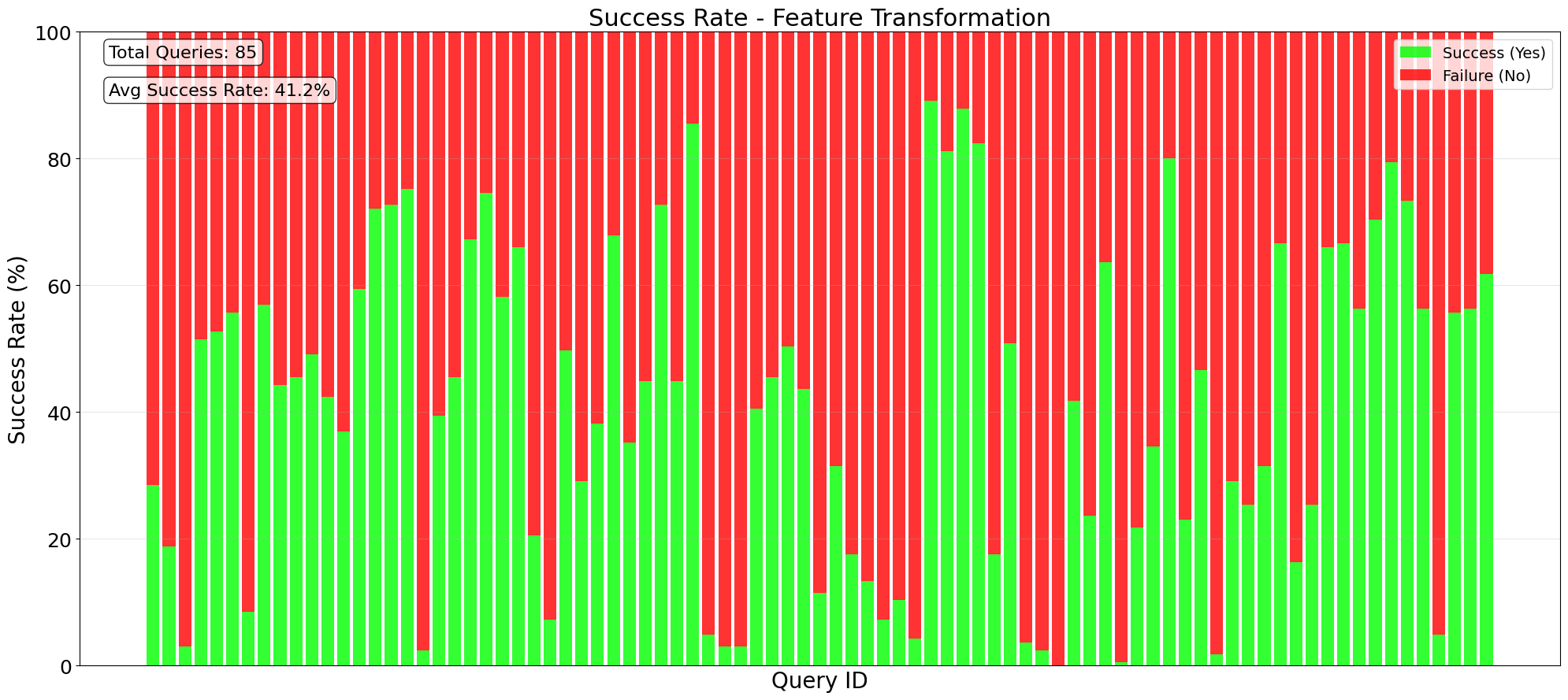}
    \caption{Success rates for each sample over several (165) attempts - Feature Transformation}
    \label{figure:06F}
\end{figure*}
\begin{figure*}[!h]
    \centering
    \includegraphics[width=0.93\linewidth]{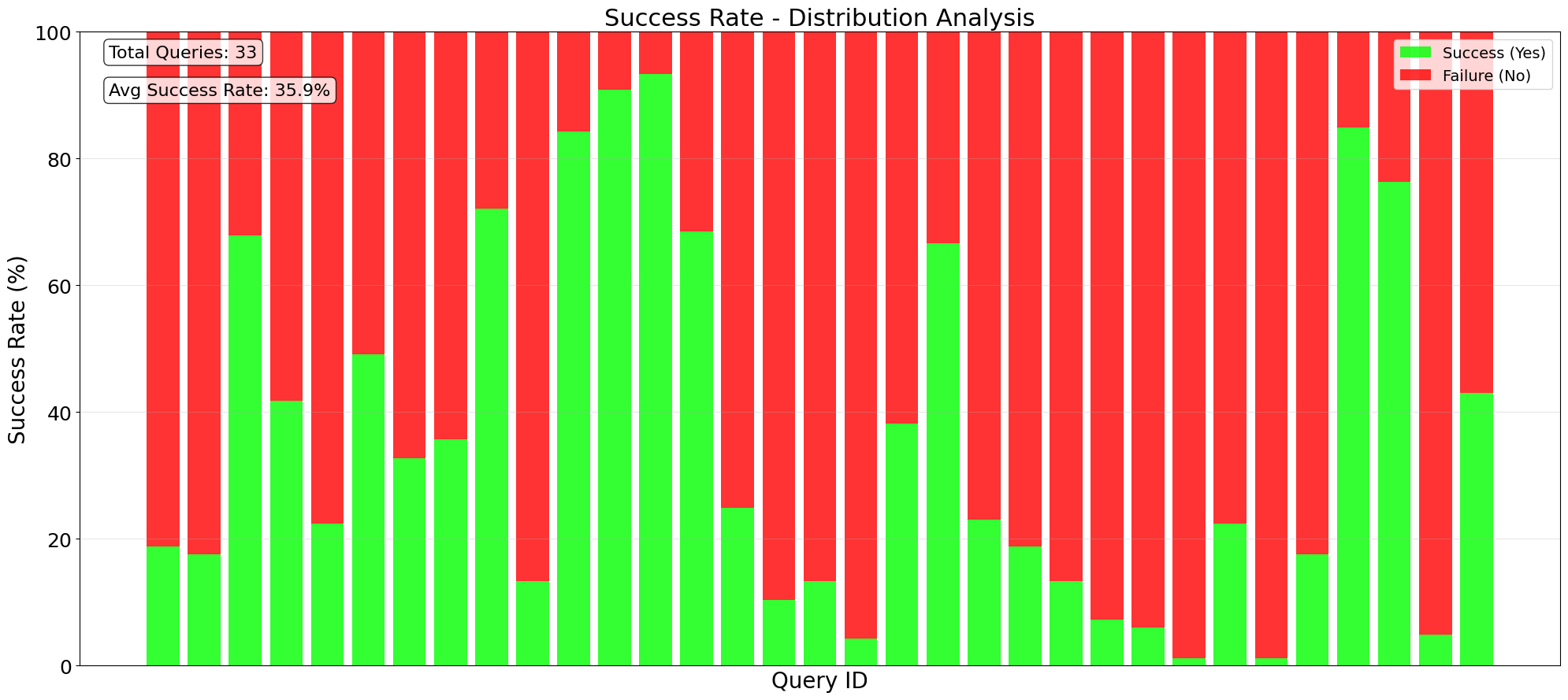}
    \caption{Success rates for each sample over several (165) attempts - Distribution Analysis}
    \label{figure:06G}
\end{figure*}
\begin{figure*}[!h]
    \centering
    \includegraphics[width=0.93\linewidth]{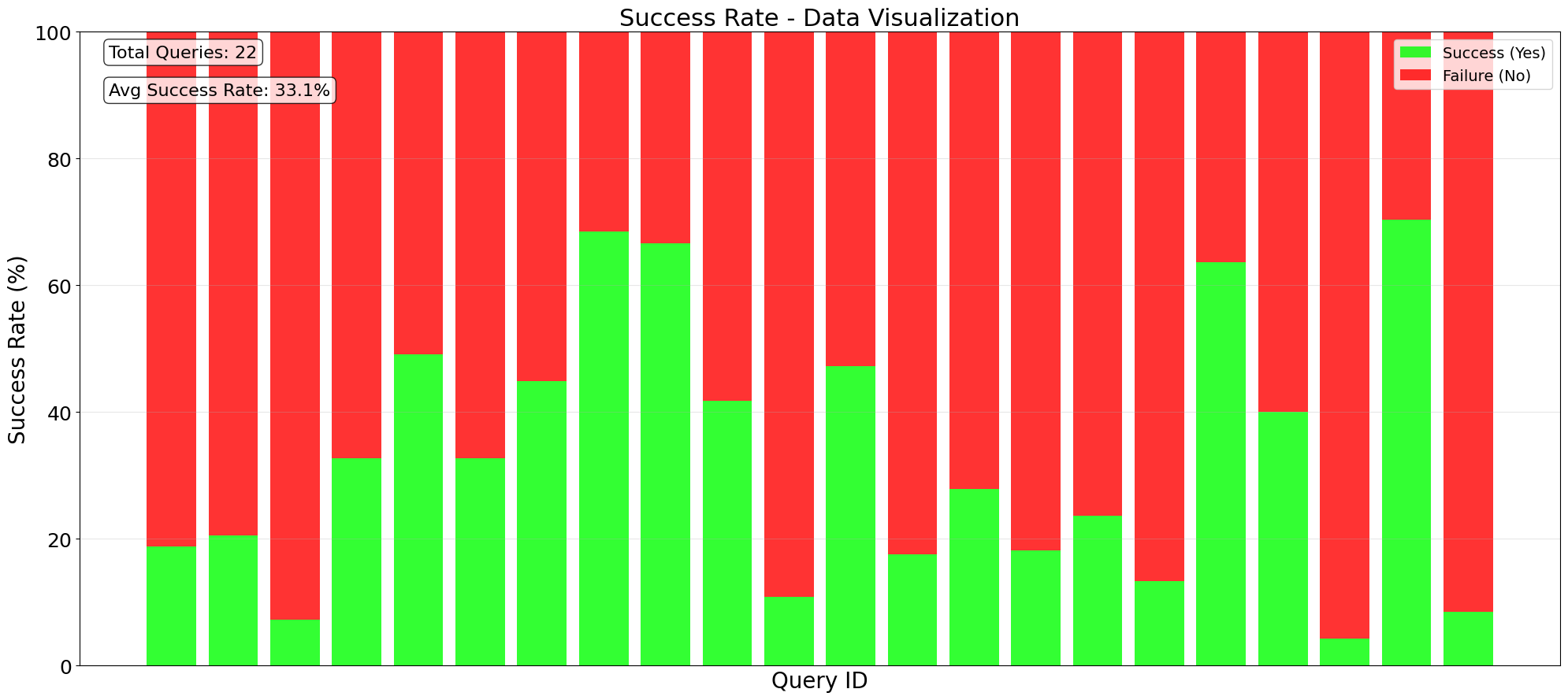}
    \caption{Success rates for each sample over several (165) attempts - Data Visualization}
    \label{figure:06H}
\end{figure*}
\newpage
\section{Accuracies over Each Task}
\label{sec:taskwiseaccuracies}
\begin{figure*}[!h]
    \centering
    \includegraphics[width=0.95\linewidth]{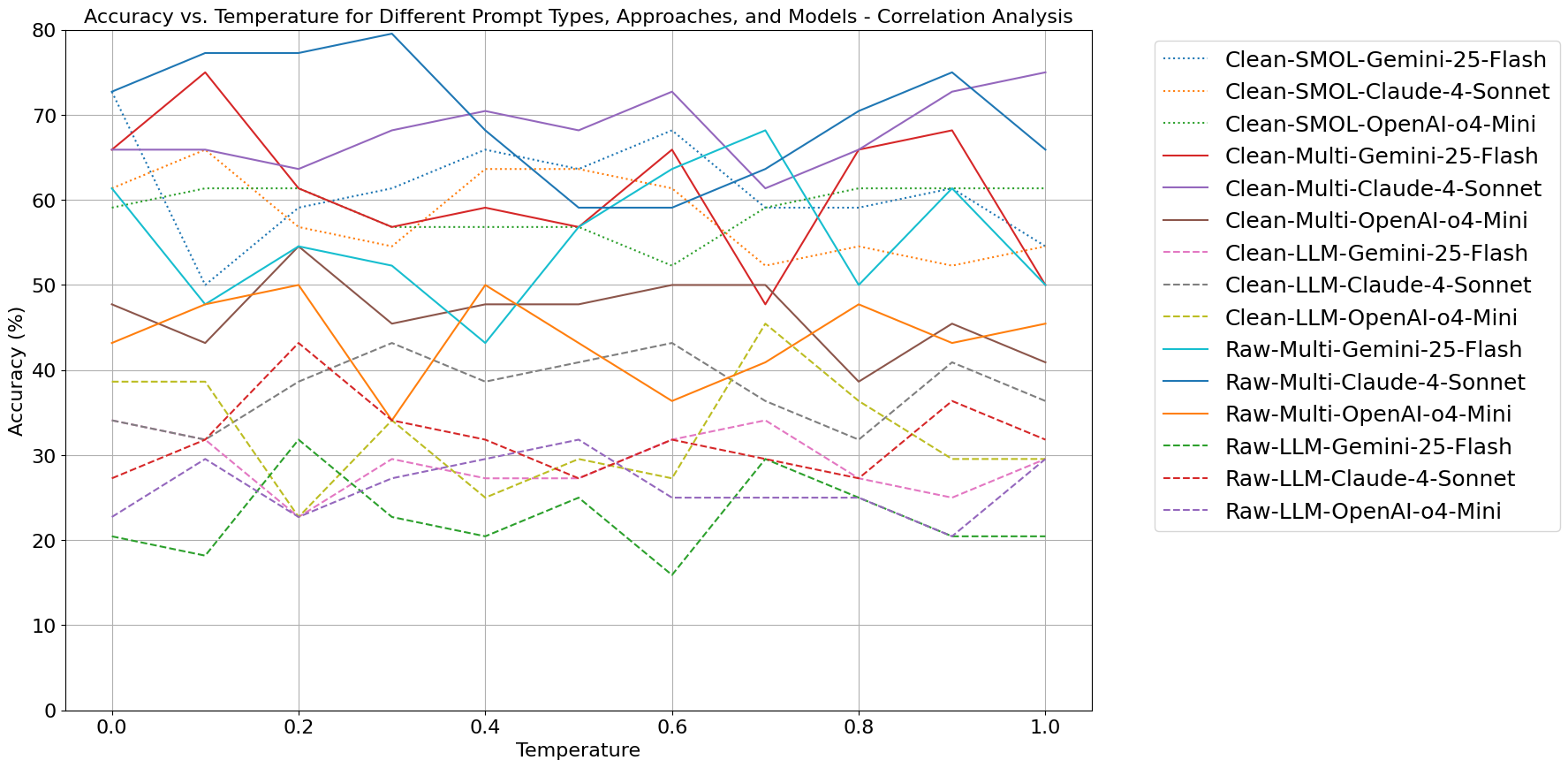}
    \caption{Results from each attempt over each temperature value used - Correlation Analysis}
    \label{figure:03A}
\end{figure*}

\begin{figure*}[!h]
    \centering
    \includegraphics[width=0.95\linewidth]{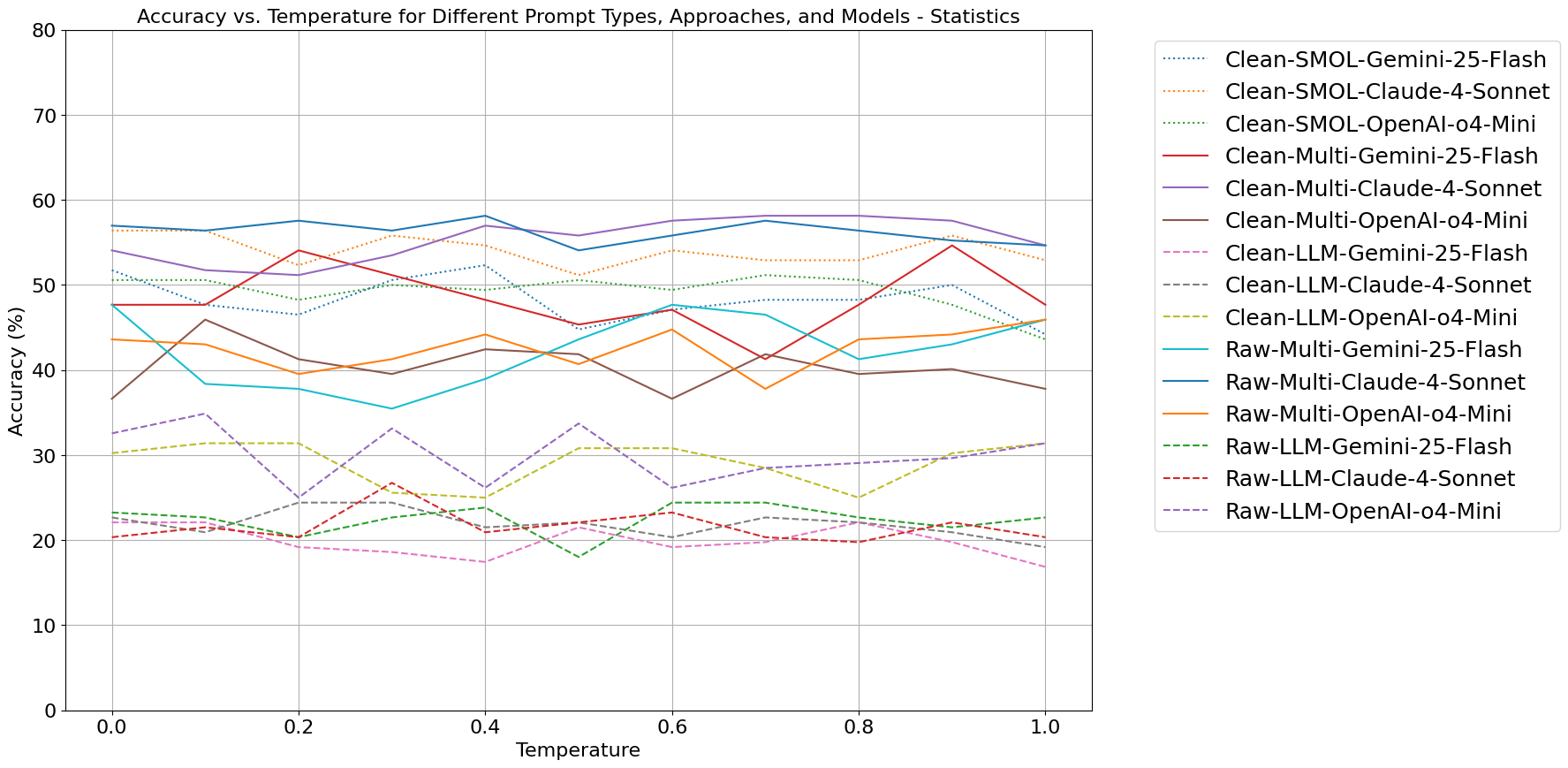}
    \caption{Results from each attempt over each temperature value used - Statistics}
    \label{figure:03B}
\end{figure*}

\begin{figure*}[!h]
    \centering
    \includegraphics[width=0.85\linewidth]{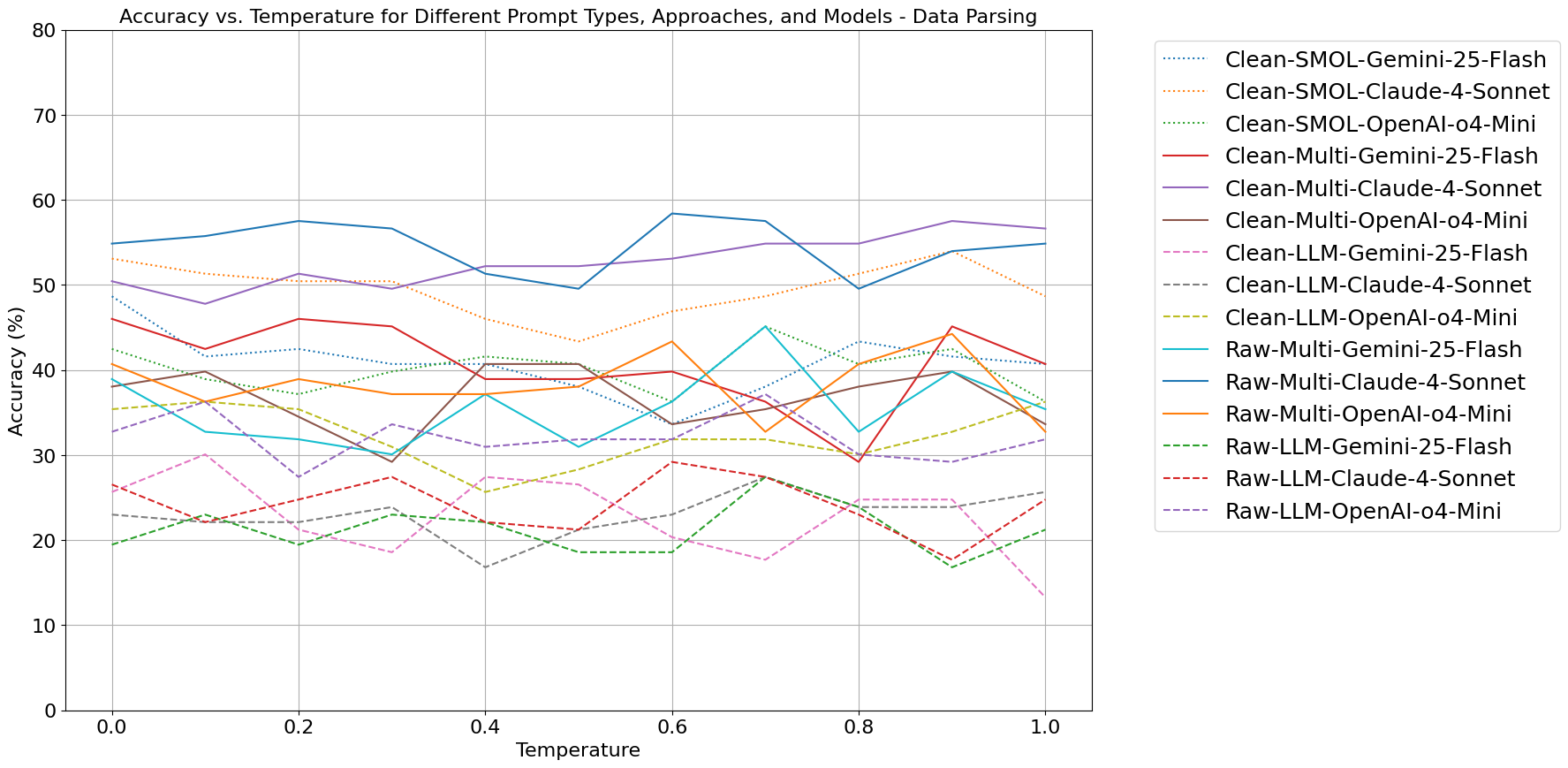}
    \caption{Results from each attempt over each temperature value used - Data Parsing}
    \label{figure:03C}
\end{figure*}

\begin{figure*}[!h]
    \centering
    \includegraphics[width=0.85\linewidth]{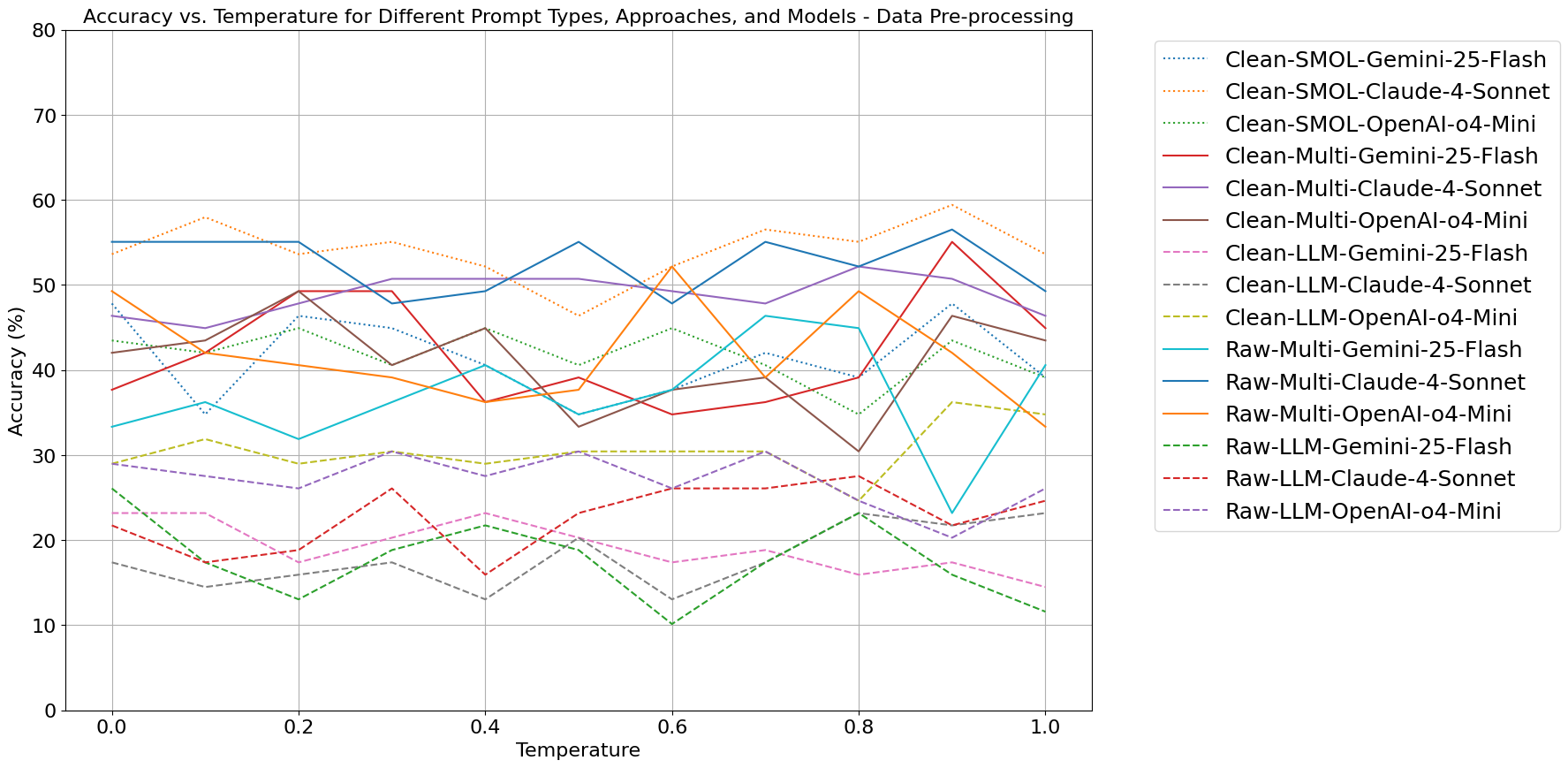}
    \caption{Results from each attempt over each temperature value used - Data Pre-processing}
    \label{figure:03D}
\end{figure*}

\begin{figure*}[!h]
    \centering
    \includegraphics[width=0.85\linewidth]{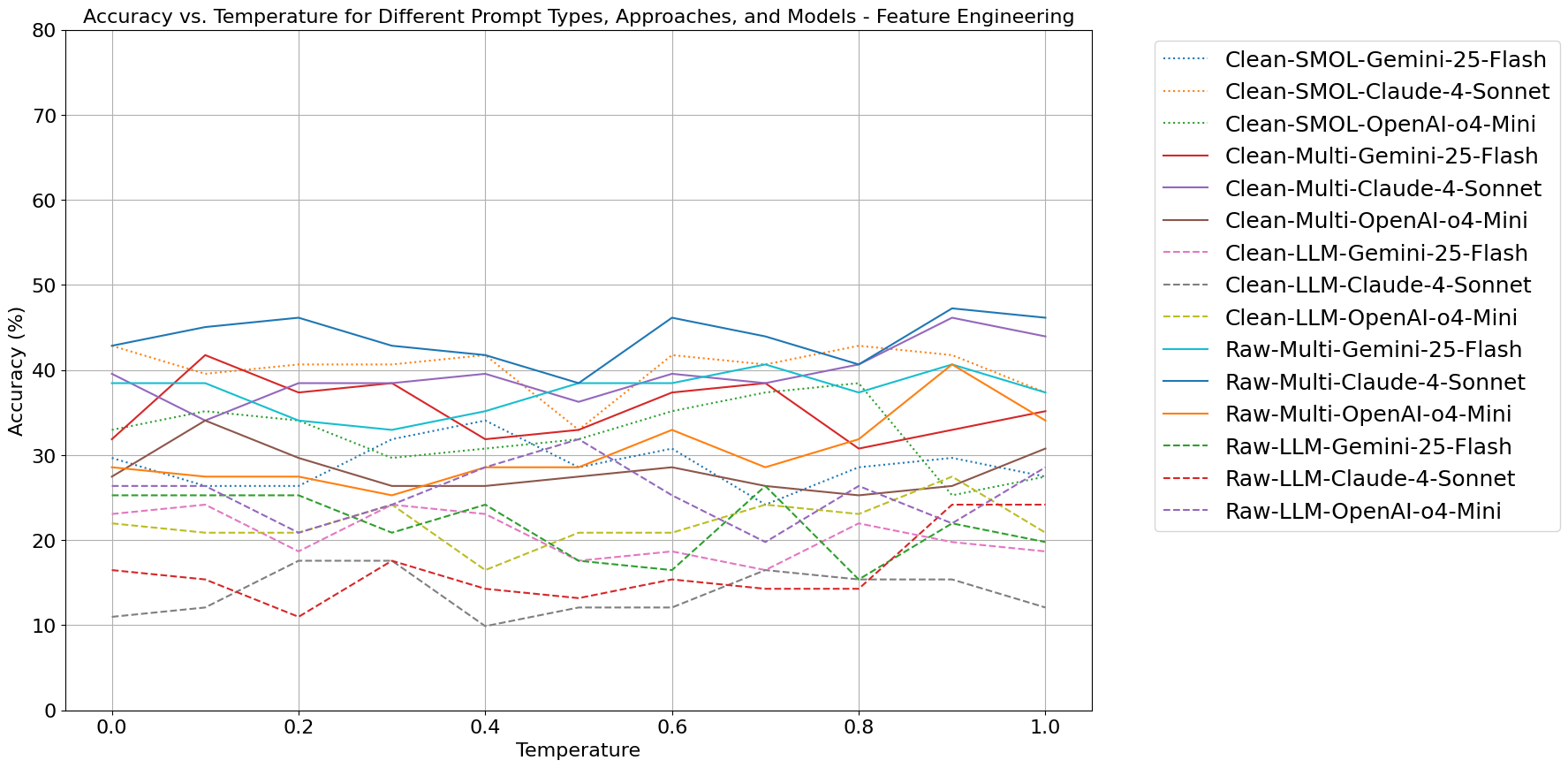}
    \caption{Results from each attempt over each temperature value used - Feature Engineering}
    \label{figure:03E}
\end{figure*}

\begin{figure*}[!h]
    \centering
    \includegraphics[width=0.85\linewidth]{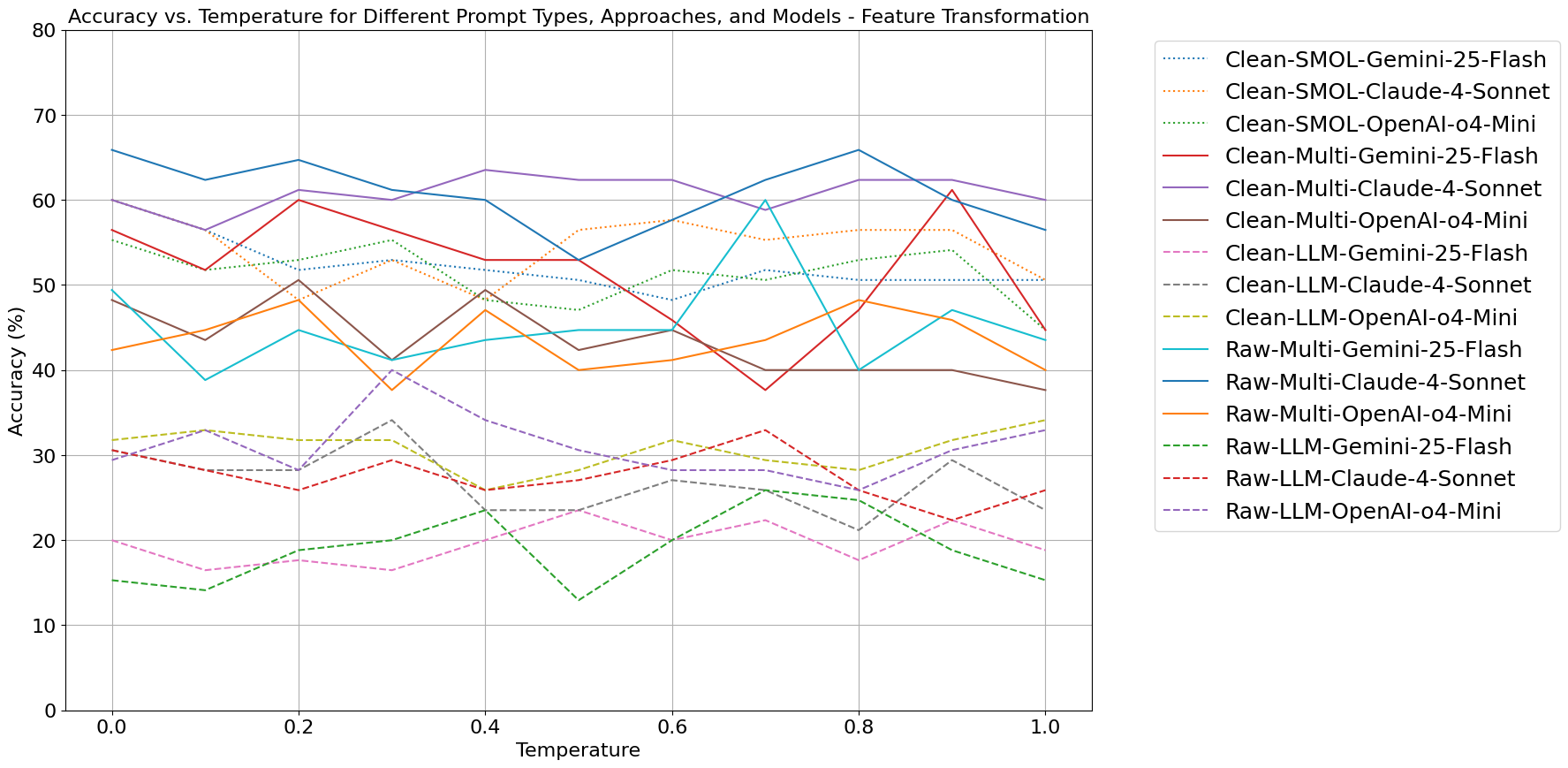}
    \caption{Results from each attempt over each temperature value used - Feature Transformation}
    \label{figure:03F}
\end{figure*}

\begin{figure*}[!h]
    \centering
    \includegraphics[width=0.85\linewidth]{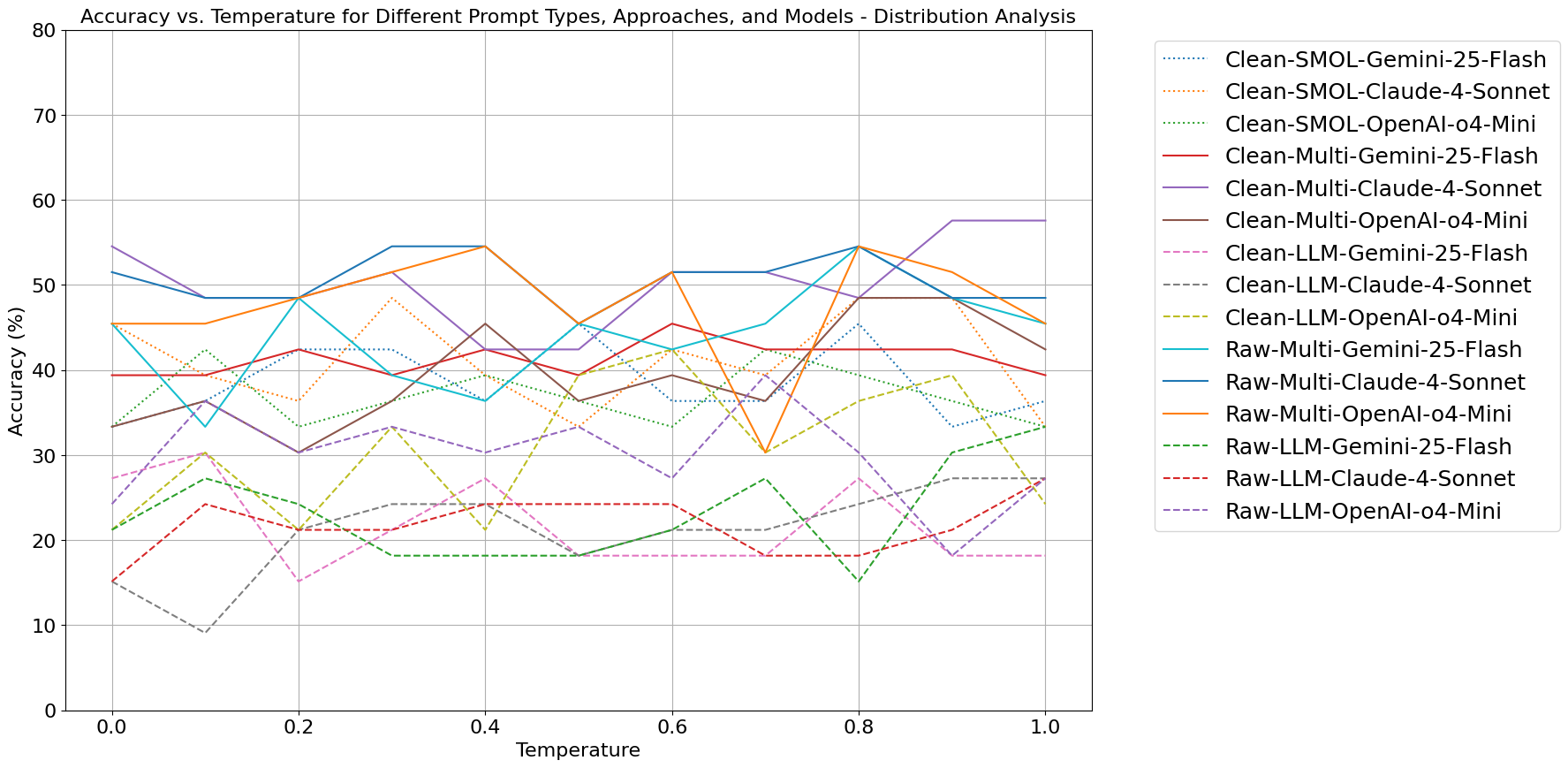}
    \caption{Results from each attempt over each temperature value used - Distribution Analysis}
    \label{figure:03G}
\end{figure*}

\begin{figure*}[!h]
    \centering
    \includegraphics[width=0.85\linewidth]{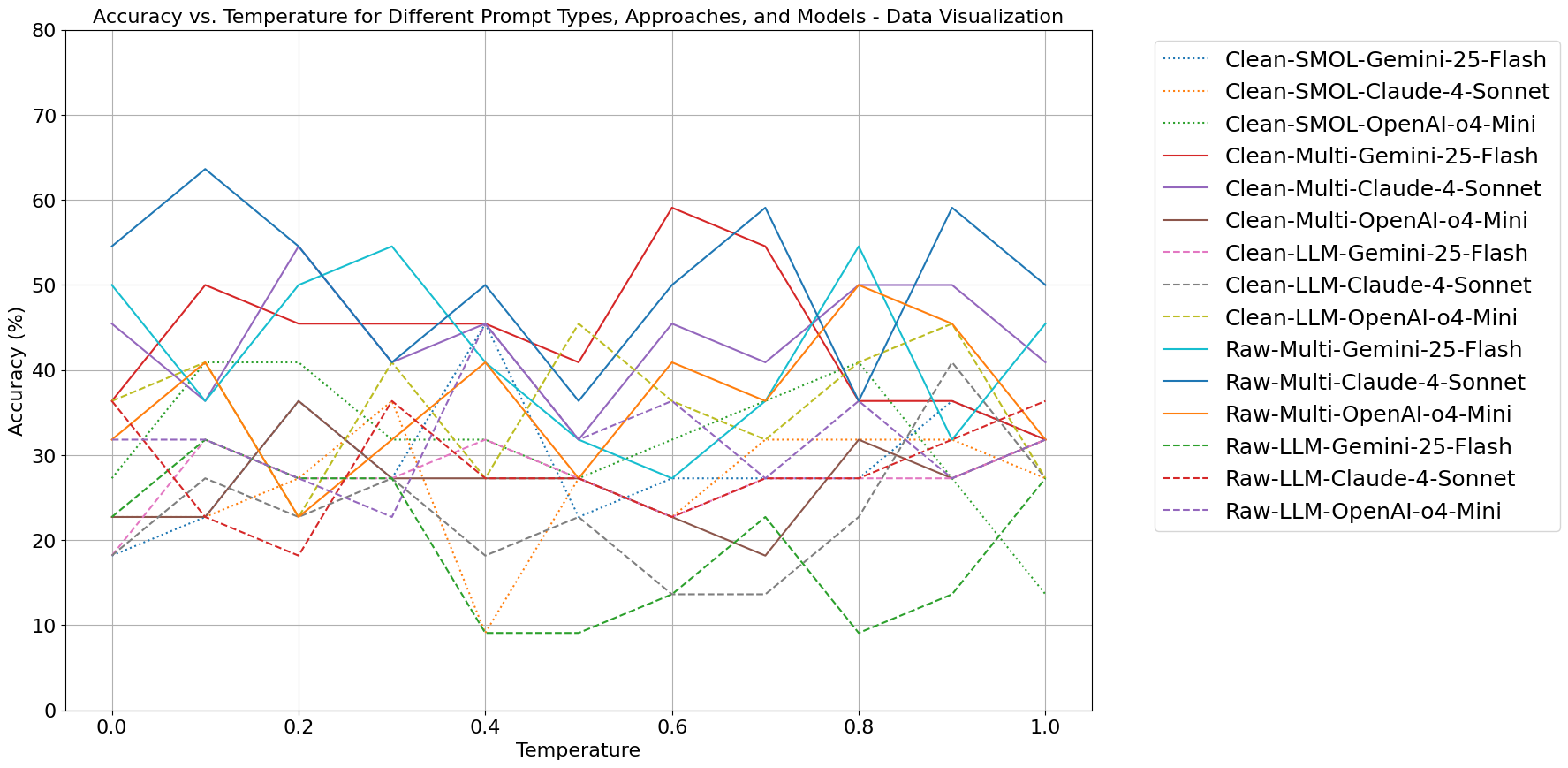}
    \caption{Results from each attempt over each temperature value used - Data Visualization}
    \label{figure:03H}
\end{figure*}

The task wise accuracies of each attempt compared to the temperature used can be seen in \autoref{figure:03A}, \autoref{figure:03B}, \autoref{figure:03C}, \autoref{figure:03D}, \autoref{figure:03E}, \autoref{figure:03F}, \autoref{figure:03G}, \autoref{figure:03H} respectively for each task category separately. Some of the tasks had less variance between the accuracies obtained through each attempt while the rest had very high variance. \textbf{This hints that certain tasks might perform more or less the same with any set of (temperature, LLM, approach, query type) while others perform clearly better with a certain set of the same features.}  This is especially important as costs incurred can vary a lot between using multi-code-cell approach or by SmolAgent compared to generating them directly by an LLM in a single step. Though the difference in costs is only a few cents per sample, it can make a difference when dealing with a large number of samples.
\newpage
\newpage
\newpage
\section{Variation in Accuracies : task wise}
\label{sec:variationinaccuracies}
\begin{figure*}[!h]
    \centering
    \includegraphics[width=0.7\linewidth]{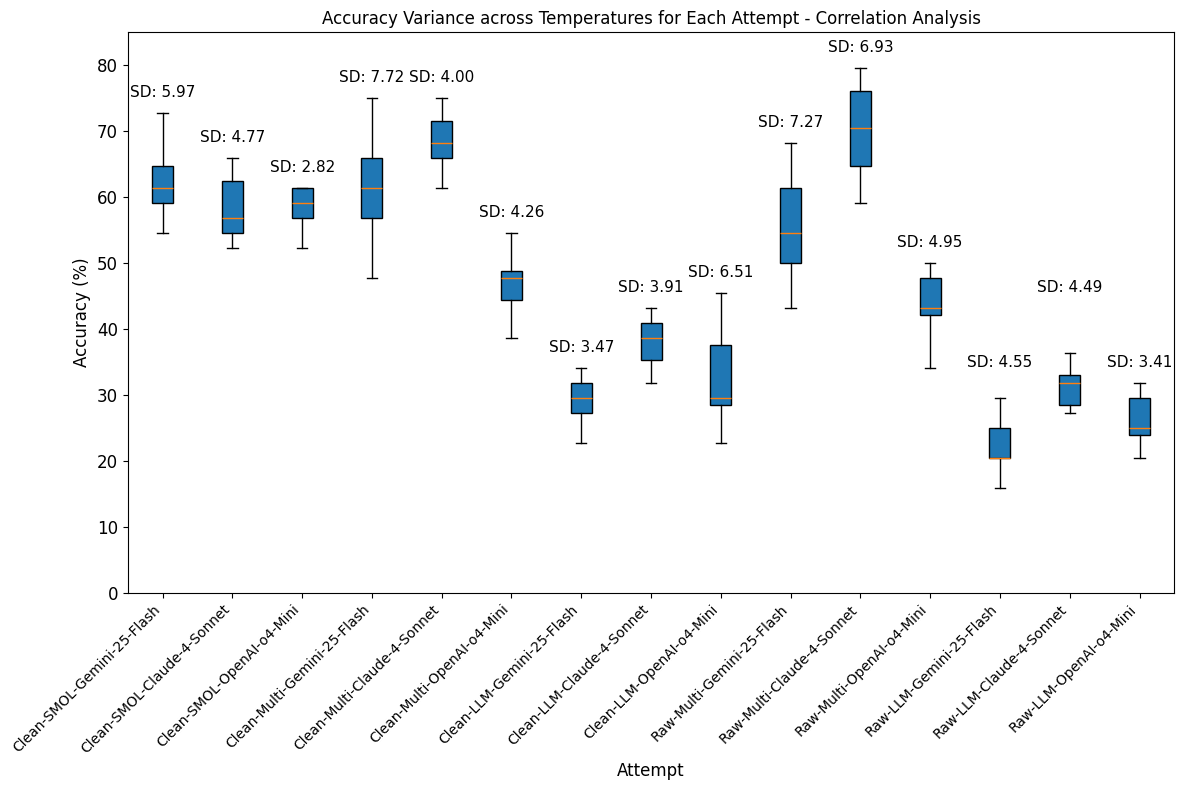}
    \caption{Variation in Accuracies through - Correlation Analysis}
    \label{figure:05A}
\end{figure*}

\begin{figure*}[!h]
    \centering
    \includegraphics[width=0.7\linewidth]{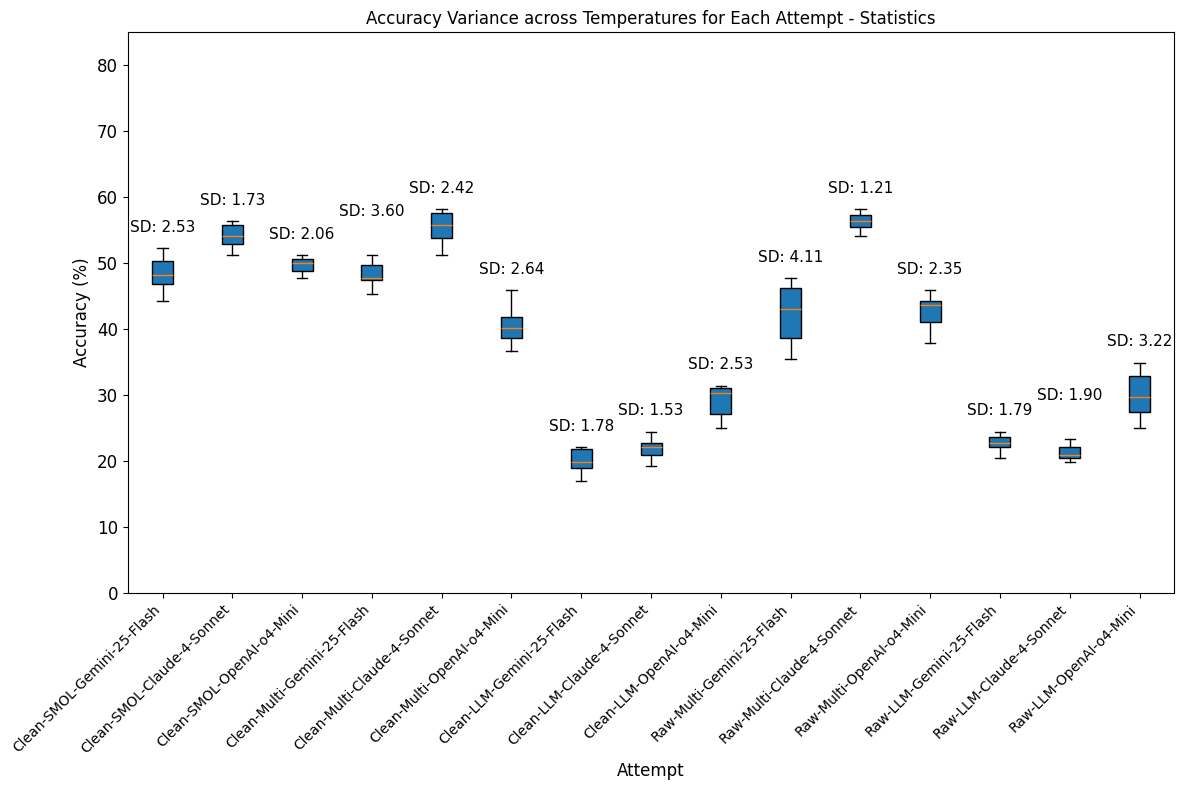}
    \caption{Variation in Accuracies through - Statistics}
    \label{figure:05B}
\end{figure*}

\begin{figure*}[!h]
    \centering
    \includegraphics[width=0.63\linewidth]{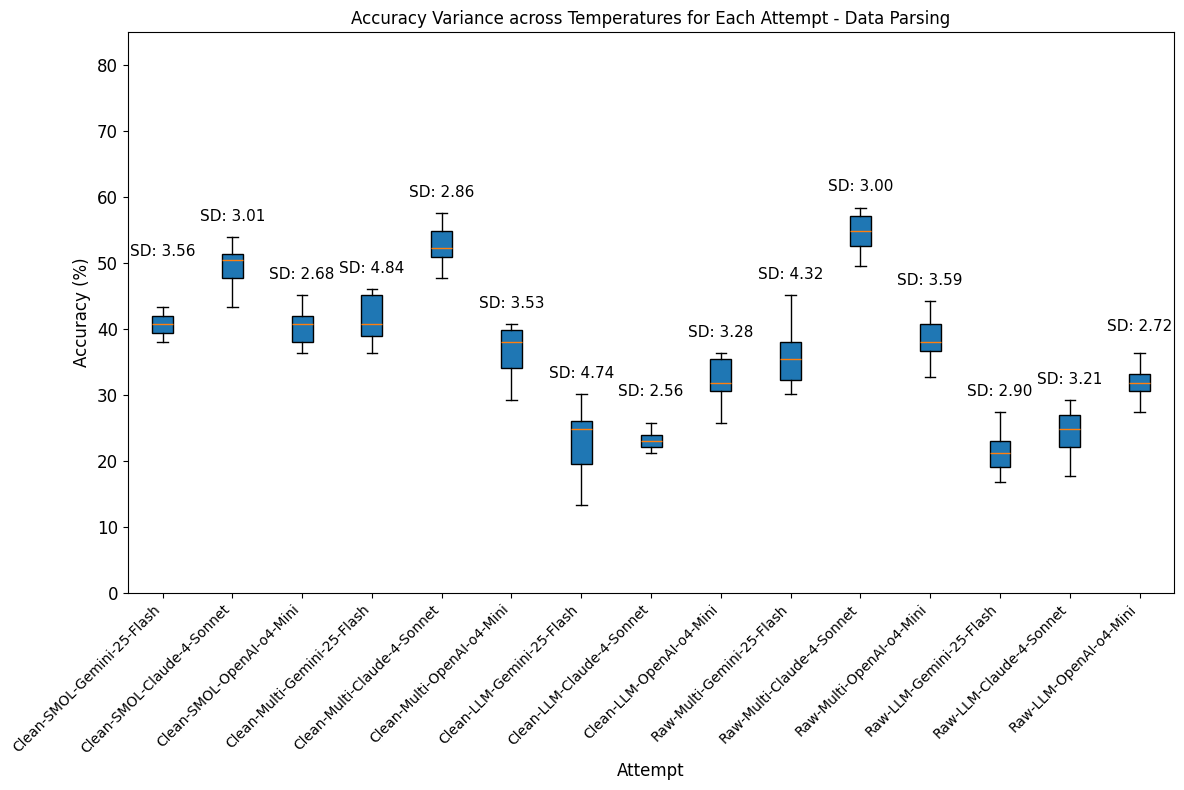}
    \caption{Variation in Accuracies through - Data Parsing}
    \label{figure:05C}
\end{figure*}

\begin{figure*}[!h]
    \centering
    \includegraphics[width=0.63\linewidth]{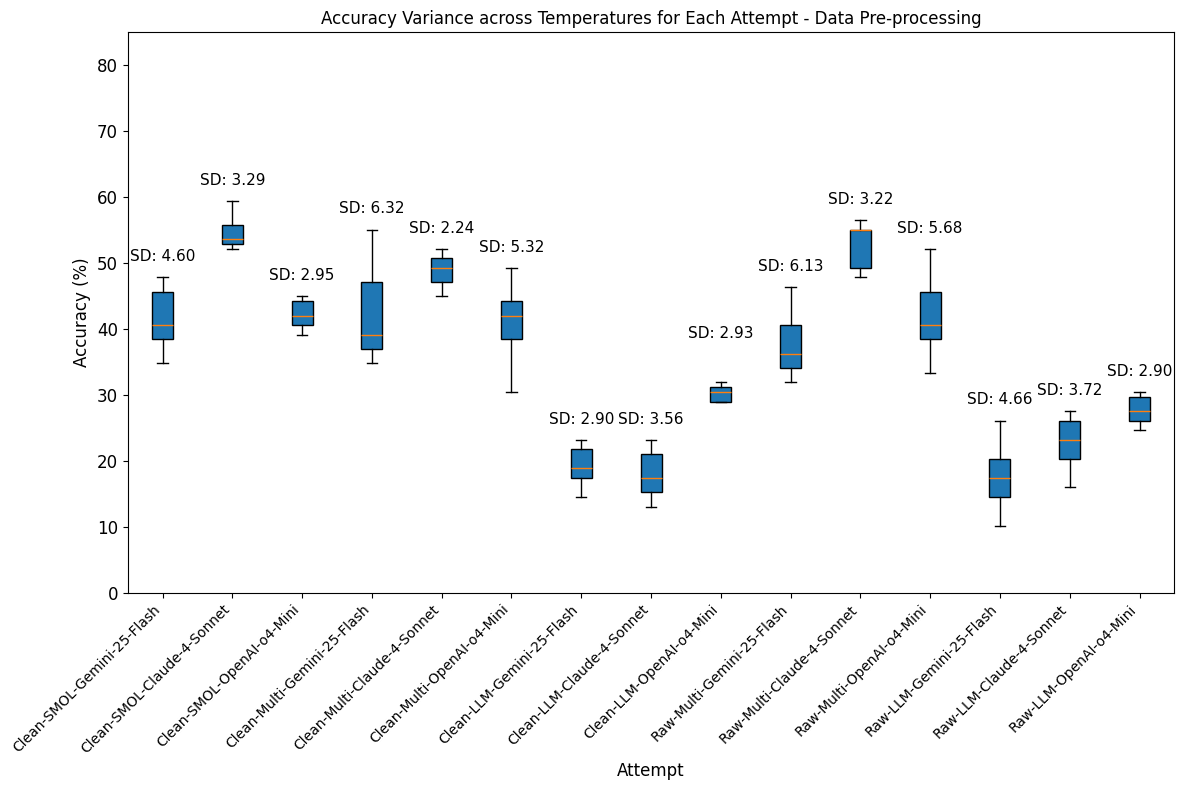}
    \caption{Variation in Accuracies through - Data Pre-processing}
    \label{figure:05D}
\end{figure*}

\begin{figure*}[!h]
    \centering
    \includegraphics[width=0.63\linewidth]{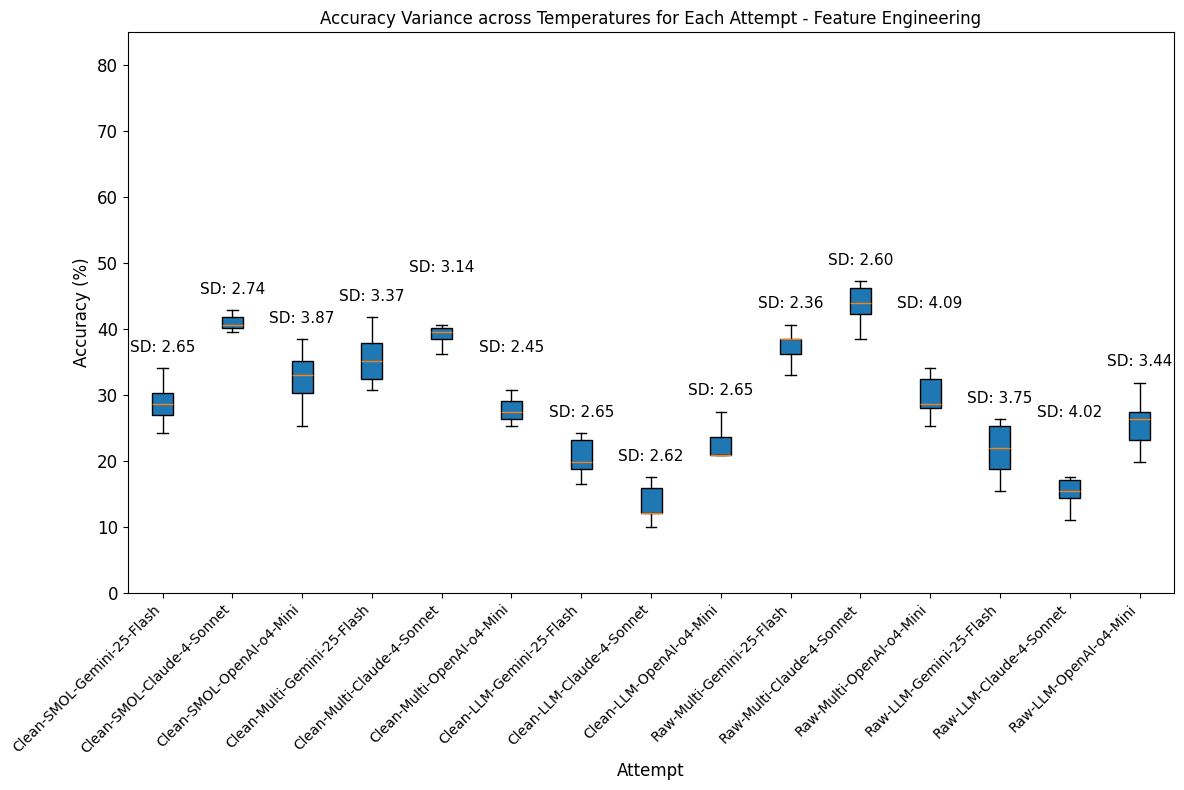}
    \caption{Variation in Accuracies through - Feature Engineering}
    \label{figure:05E}
\end{figure*}

\begin{figure*}[!h]
    \centering
    \includegraphics[width=0.63\linewidth]{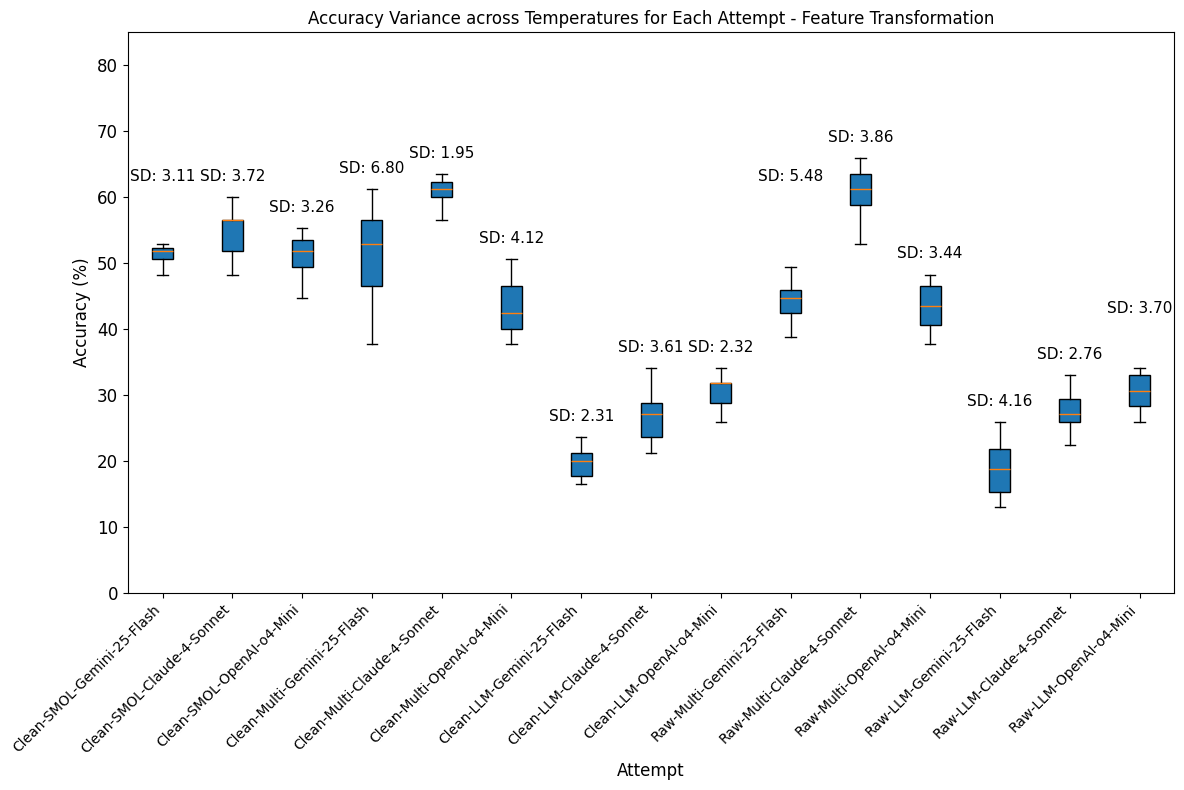}
    \caption{Variation in Accuracies through - Feature Transformation}
    \label{figure:05F}
\end{figure*}

\begin{figure*}[!h]
    \centering
    \includegraphics[width=0.63\linewidth]{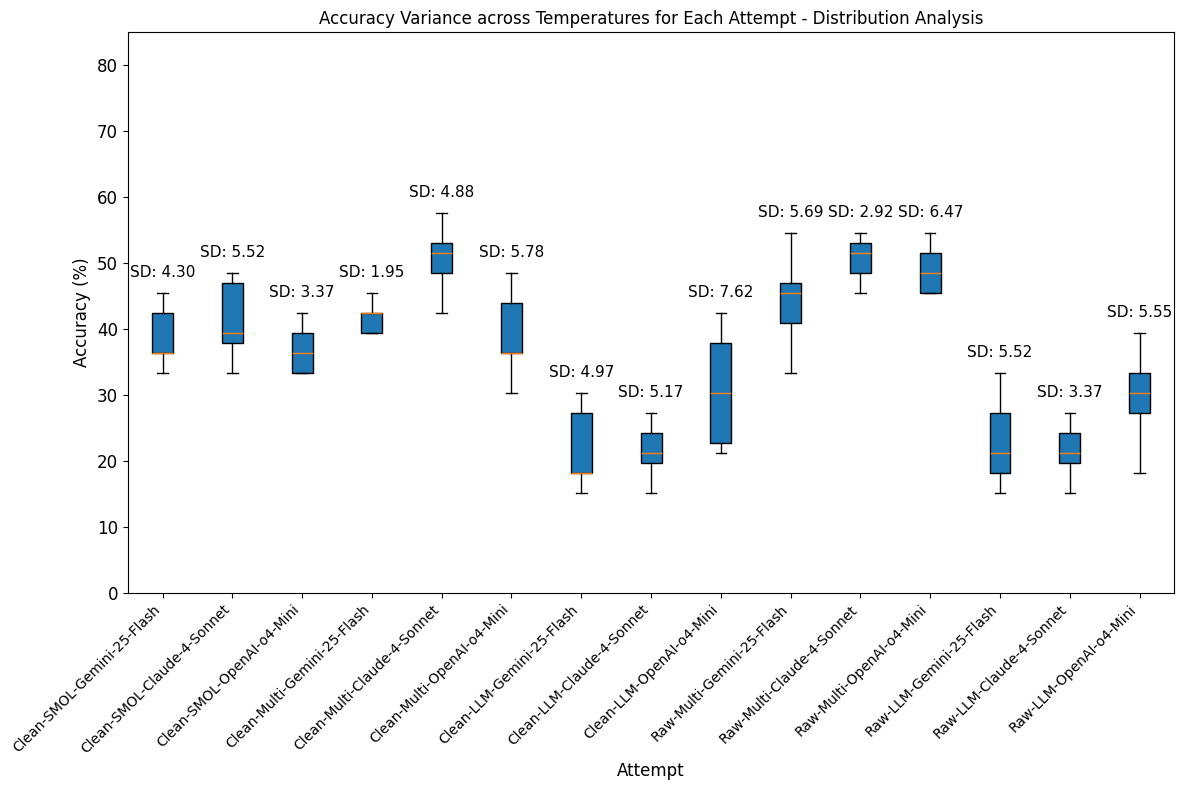}
    \caption{Variation in Accuracies through - Distribution Analysis}
    \label{figure:05G}
\end{figure*}

\begin{figure*}[!h]
    \centering
    \includegraphics[width=0.63\linewidth]{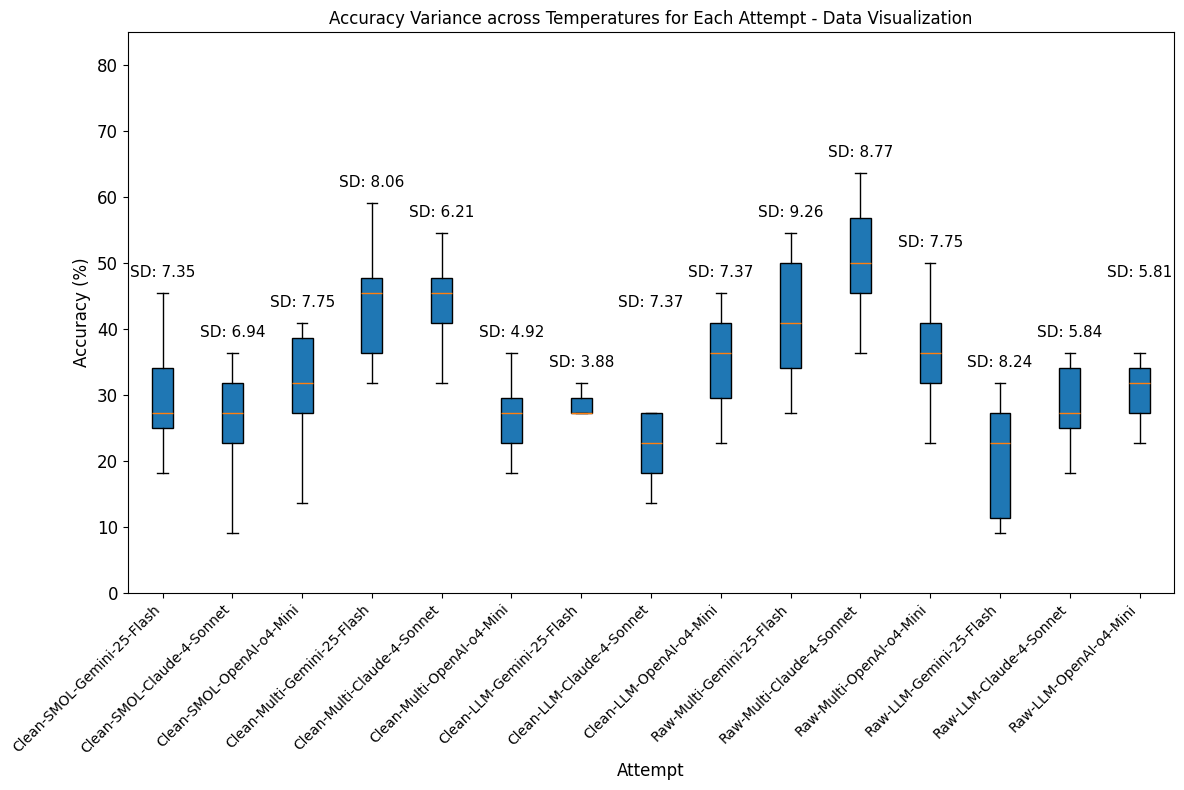}
    \caption{Variation in Accuracies through - Data Visualization}
    \label{figure:05H}
\end{figure*}

The task wise variations in accuracies of each attempt can be seen in \autoref{figure:05A}, \autoref{figure:05B}, \autoref{figure:05C}, \autoref{figure:05D}, \autoref{figure:05E}, \autoref{figure:05F}, \autoref{figure:05G}, \autoref{figure:05H} respectively for each task category separately. \textbf{Some task categories are less sensitive to temperature, while other tasks are more sensitive to changes in temperature.} The accuracy ranges for each task types can be seen to vary between each task category i.e some LLMs and approaches perform better in some task categories while some other LLM and approach might be better at other task categories. Hence, \textbf{Model and approach routing could be implemented by identifying the task category based on the input query}.
\newpage
\newpage
\newpage
\section{Error Analysis : Approach wise}
\label{sec:errorcounts}
\begin{figure*}[!h]
    \centering
    \includegraphics[width=0.98\linewidth]{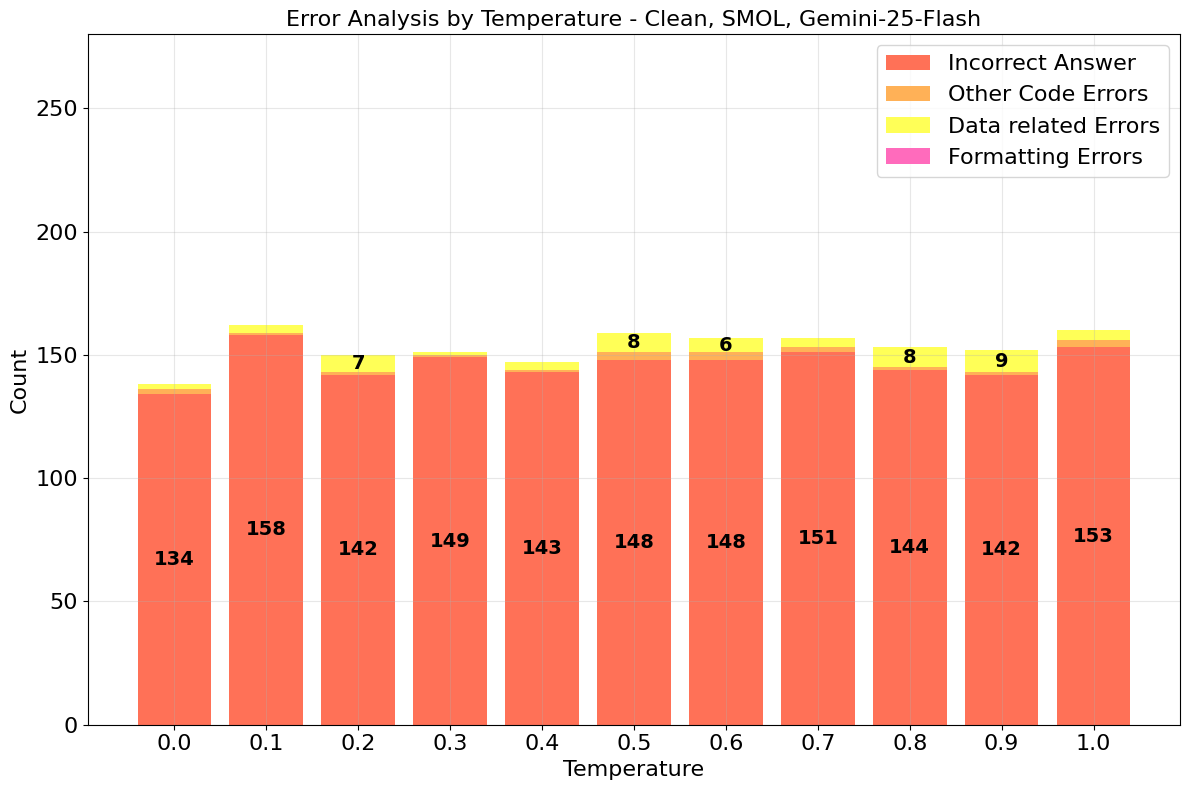}
    \caption{Incorrect response cause distribution - }
    \label{figure:08A}
\end{figure*}
\begin{figure*}[!h]
    \centering
    \includegraphics[width=1\linewidth]{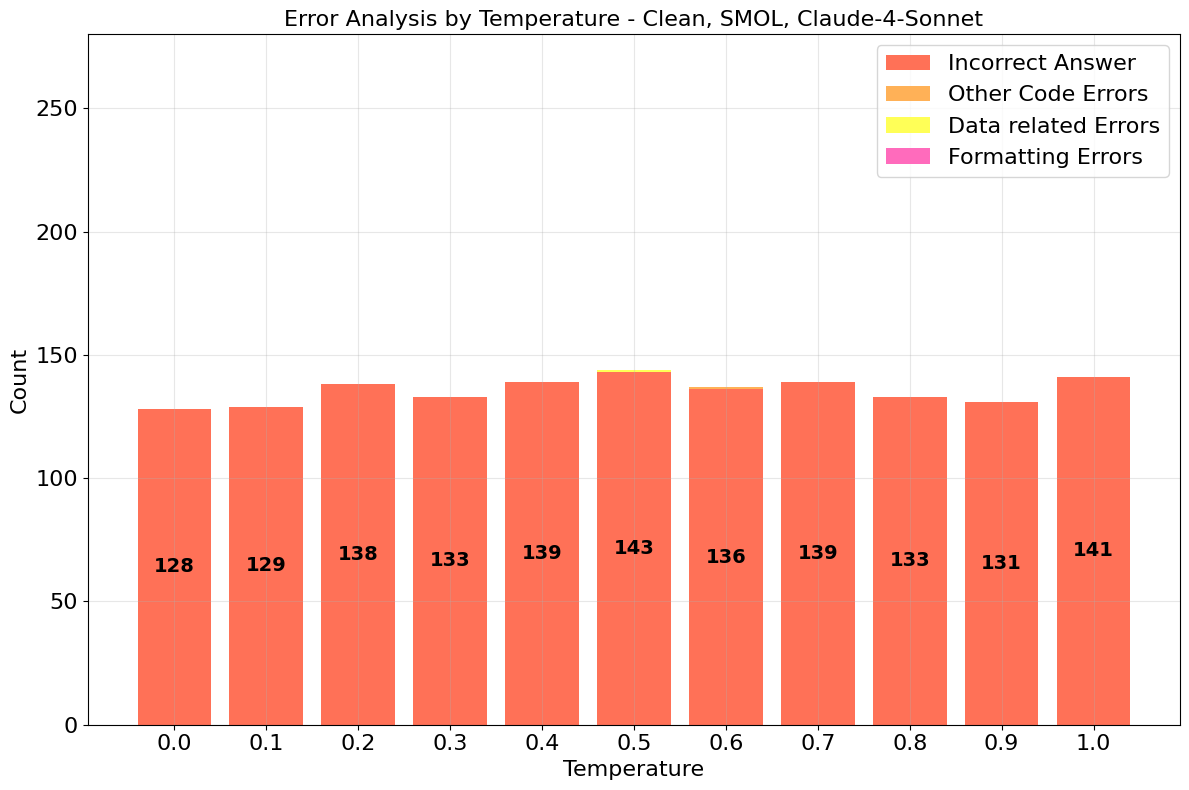}
    \caption{Incorrect response cause distribution - }
    \label{figure:08B}
\end{figure*}
\newpage
\begin{figure*}[!h]
    \centering
    \includegraphics[width=1\linewidth]{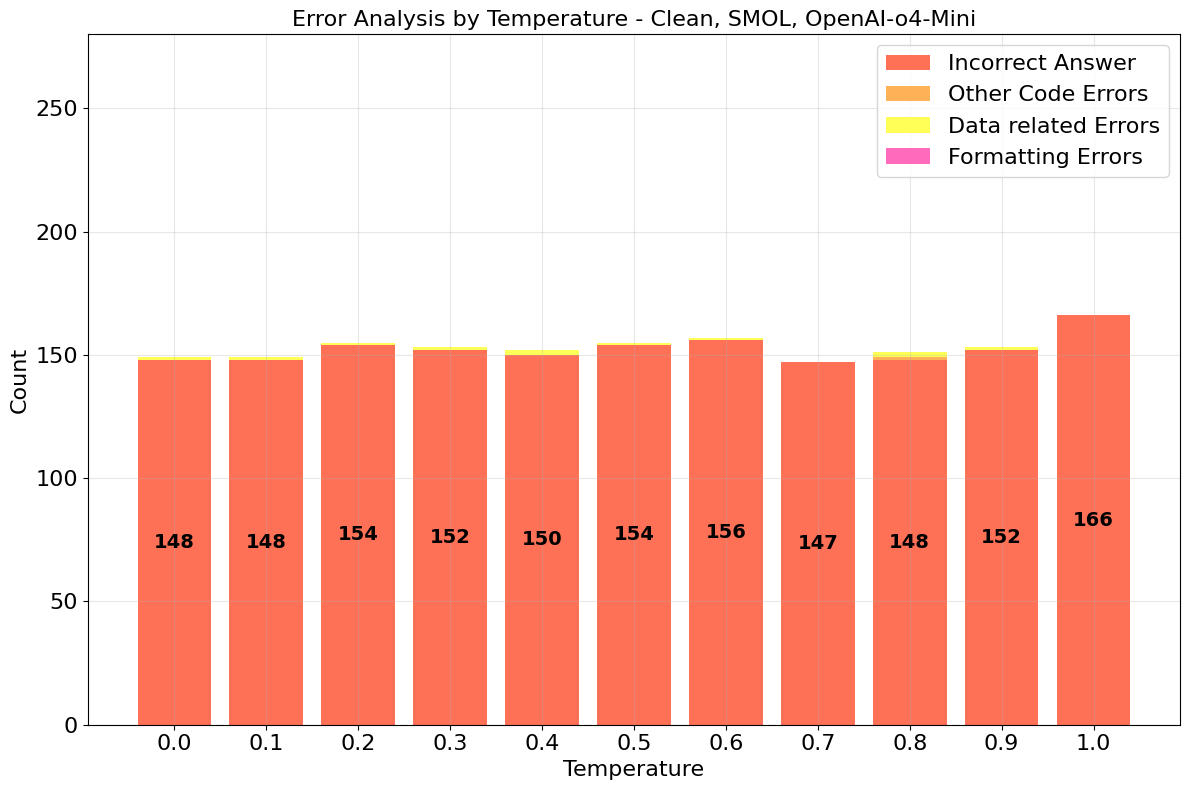}
    \caption{Incorrect response cause distribution - }
    \label{figure:08C}
\end{figure*}
\begin{figure*}[!h]
    \centering
    \includegraphics[width=1\linewidth]{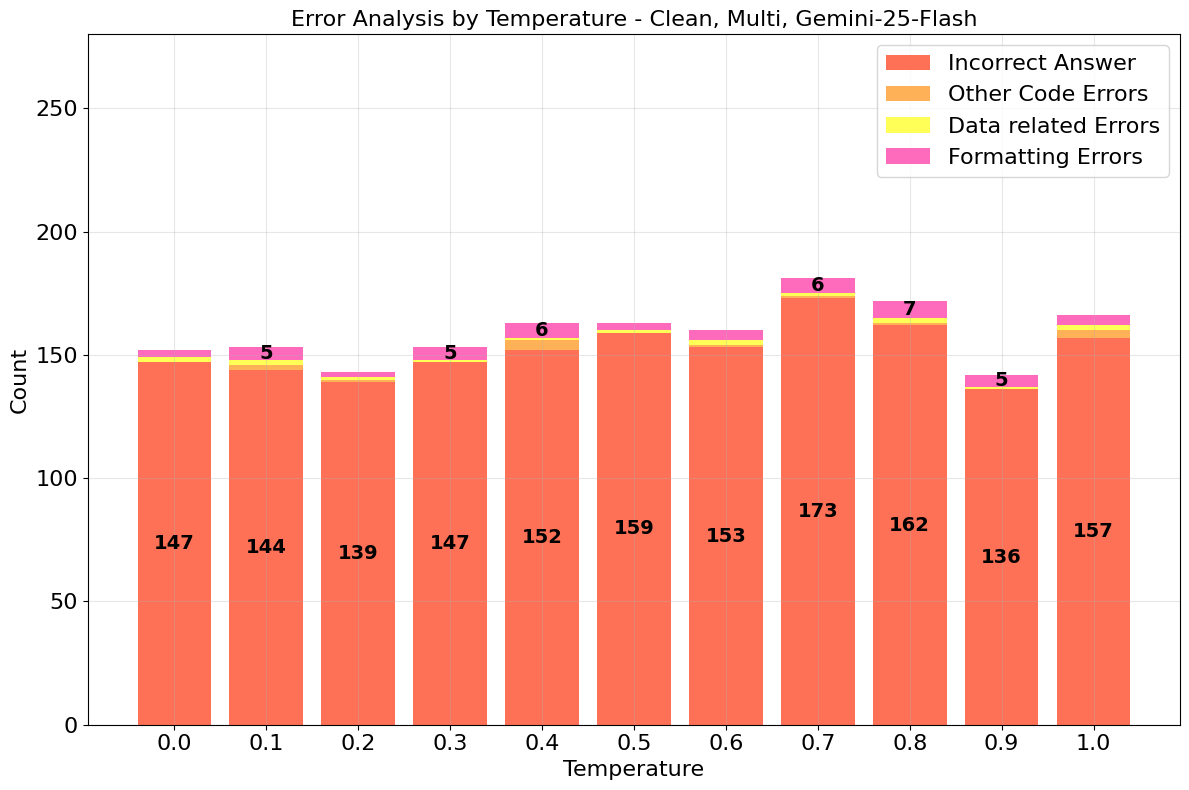}
    \caption{Incorrect response cause distribution - }
    \label{figure:08D}
\end{figure*}
\newpage
\begin{figure*}[!h]
    \centering
    \includegraphics[width=1\linewidth]{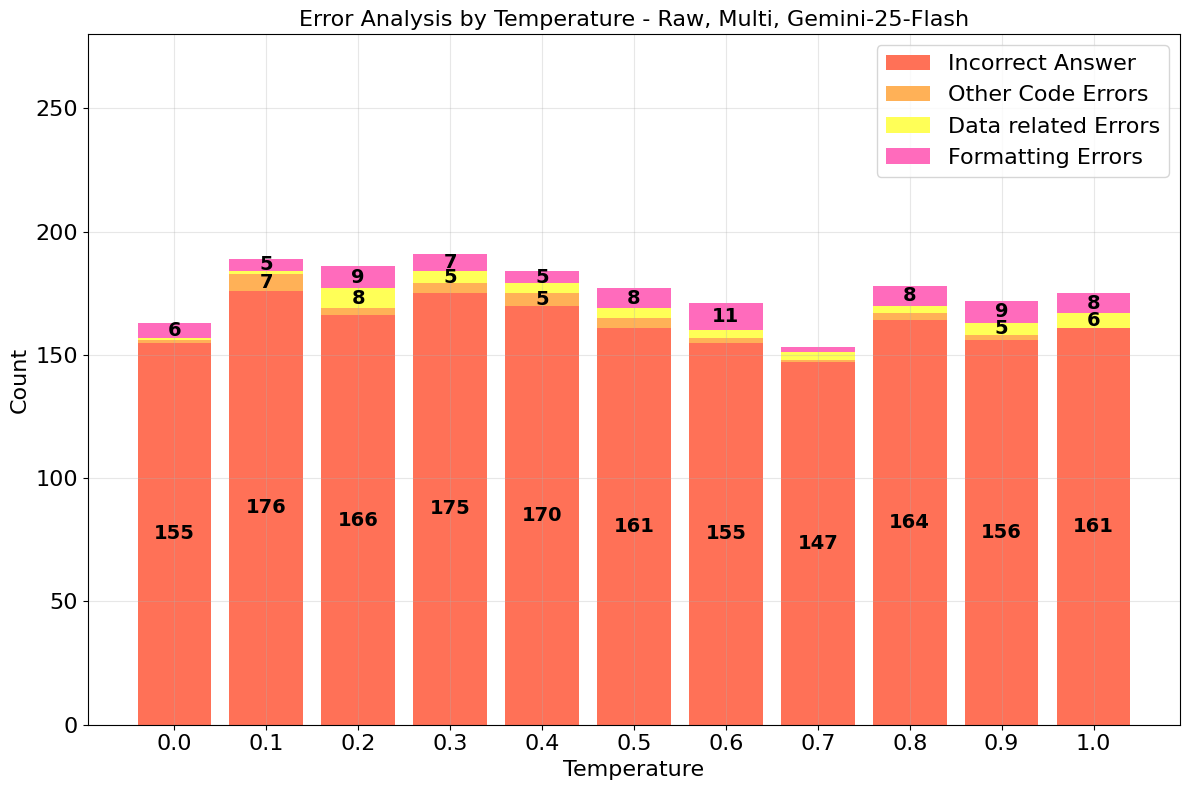}
    \caption{Incorrect response cause distribution - }
    \label{figure:08E}
\end{figure*}
\begin{figure*}[!h]
    \centering
    \includegraphics[width=1\linewidth]{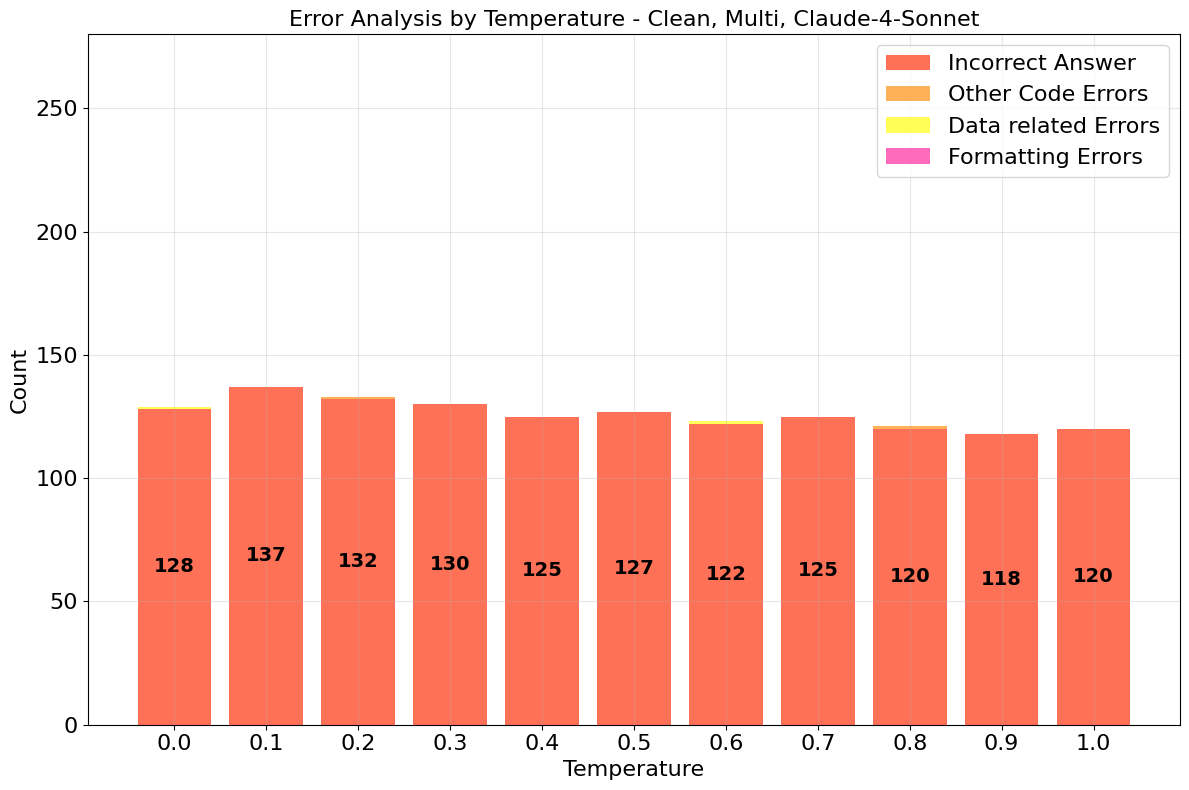}
    \caption{Incorrect response cause distribution - }
    \label{figure:08F}
\end{figure*}
\newpage
\begin{figure*}[!h]
    \centering
    \includegraphics[width=1\linewidth]{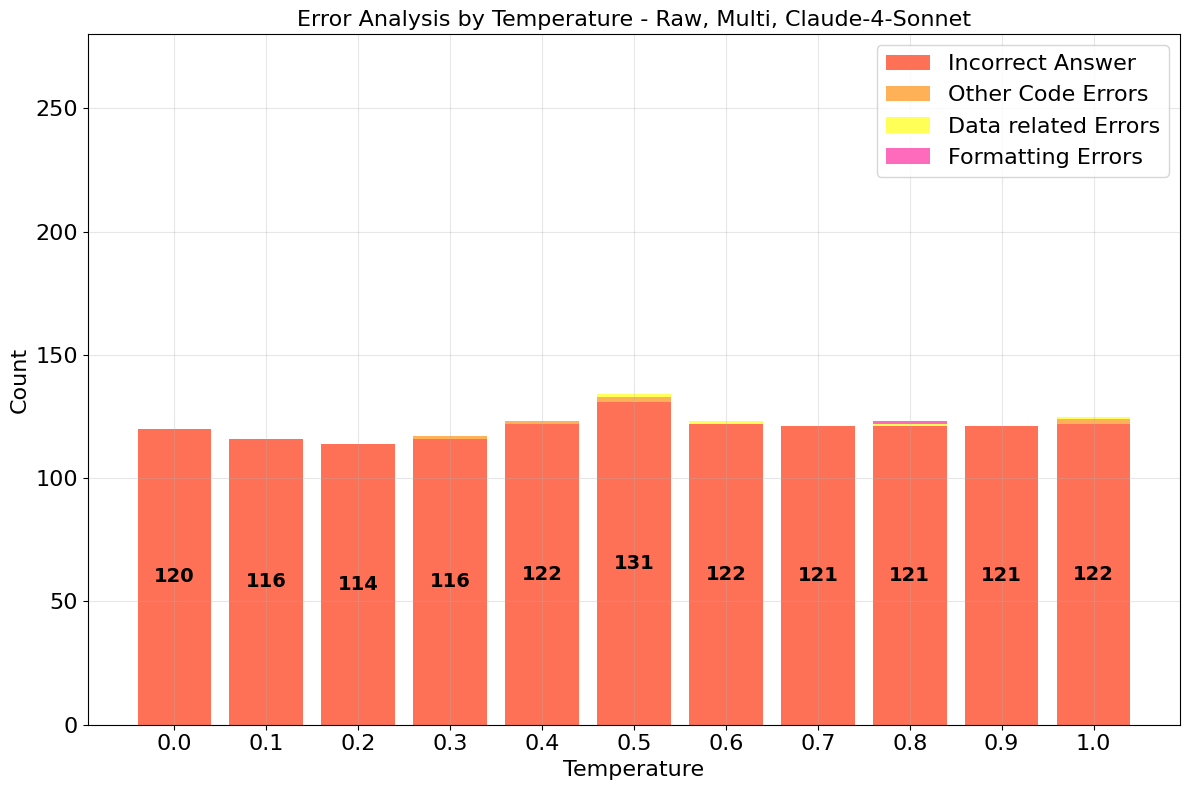}
    \caption{Incorrect response cause distribution - }
    \label{figure:08G}
\end{figure*}
\begin{figure*}[!h]
    \centering
    \includegraphics[width=1\linewidth]{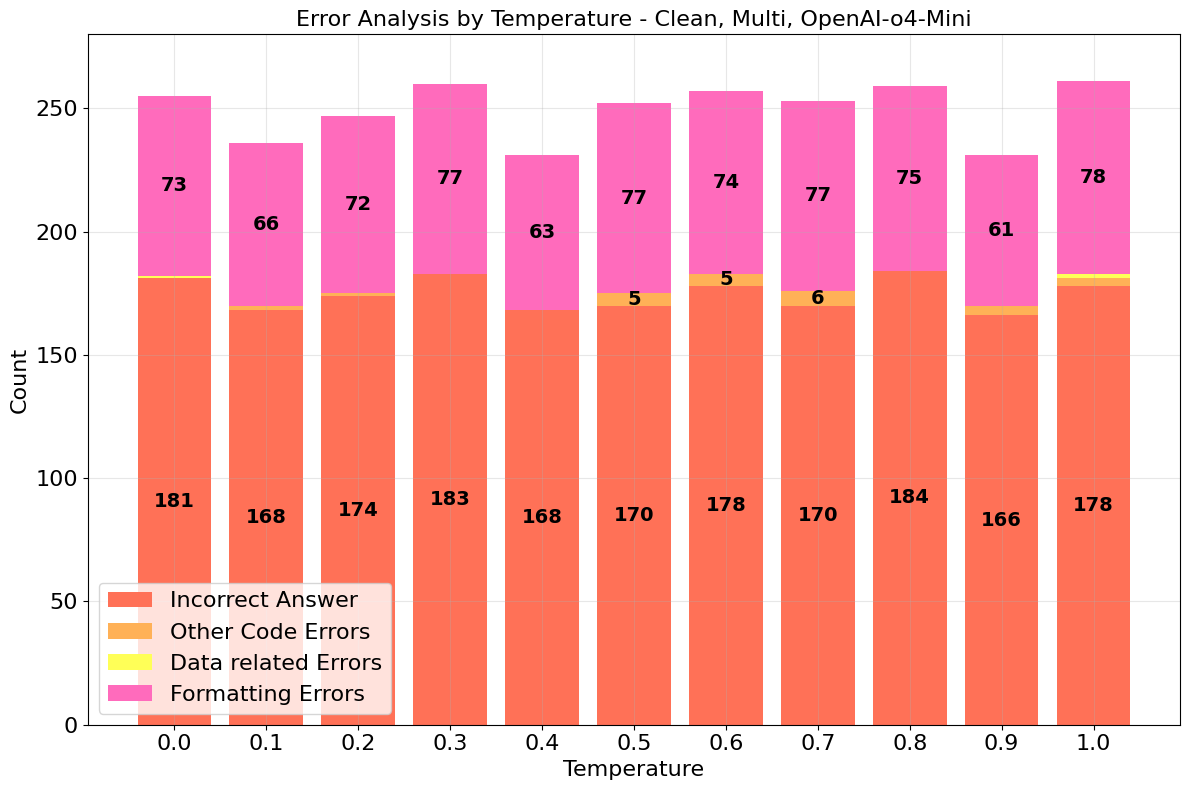}
    \caption{Incorrect response cause distribution - }
    \label{figure:08H}
\end{figure*}
\newpage
\begin{figure*}[!h]
    \centering
    \includegraphics[width=1\linewidth]{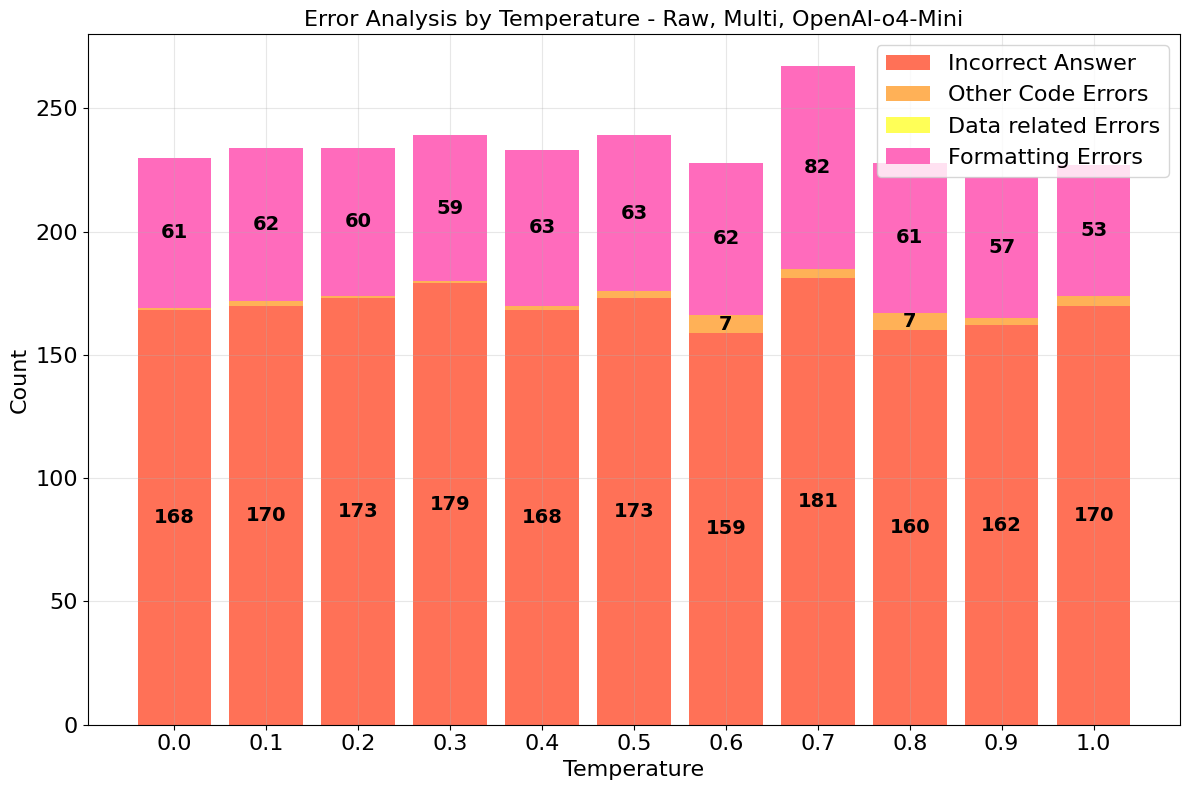}
    \caption{Incorrect response cause distribution - }
    \label{figure:08I}
\end{figure*}
\begin{figure*}[!h]
    \centering
    \includegraphics[width=1\linewidth]{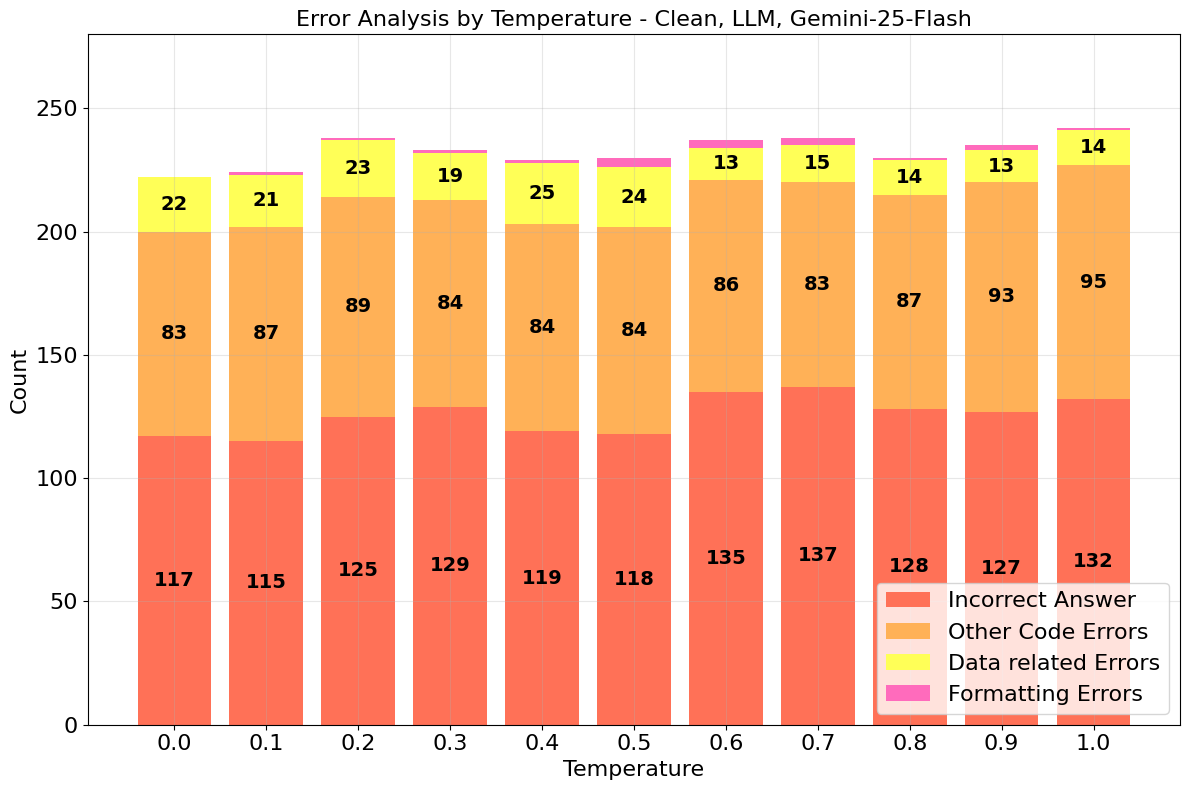}
    \caption{Incorrect response cause distribution - }
    \label{figure:08J}
\end{figure*}
\newpage
\begin{figure*}[!h]
    \centering
    \includegraphics[width=1\linewidth]{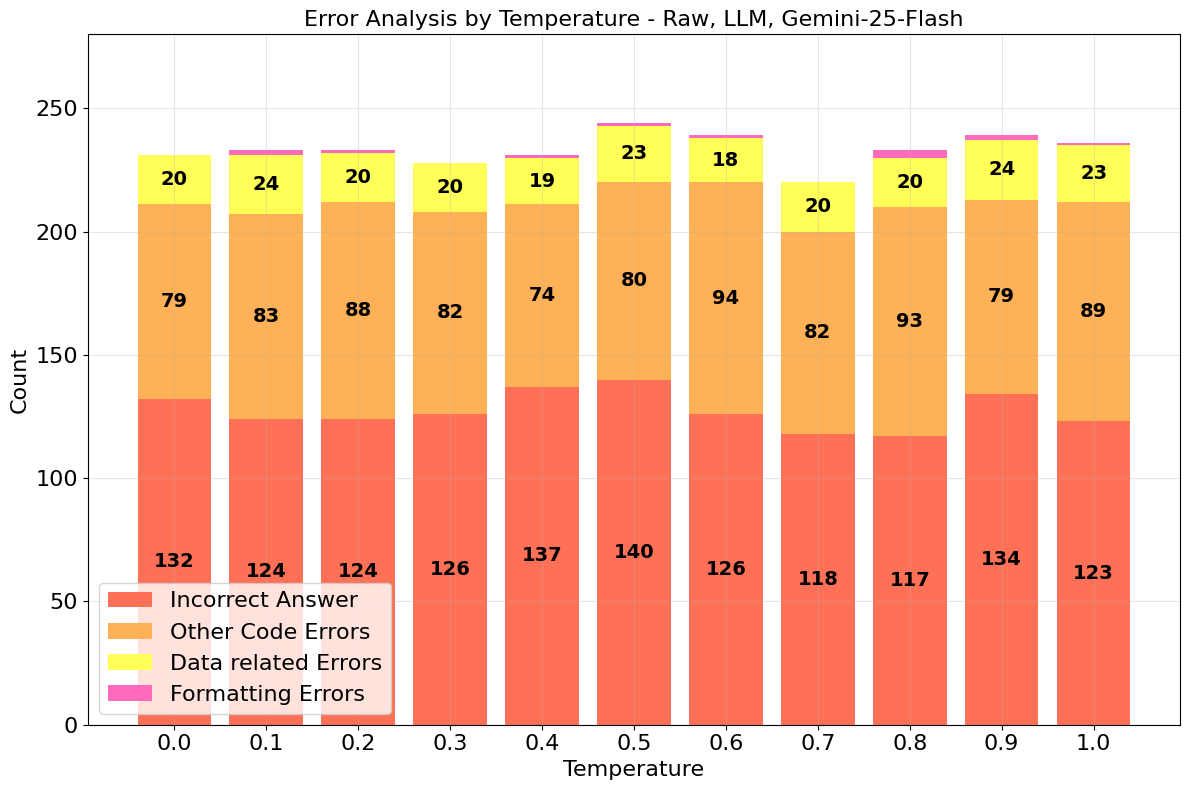}
    \caption{Incorrect response cause distribution - }
    \label{figure:08K}
\end{figure*}
\begin{figure*}[!h]
    \centering
    \includegraphics[width=1\linewidth]{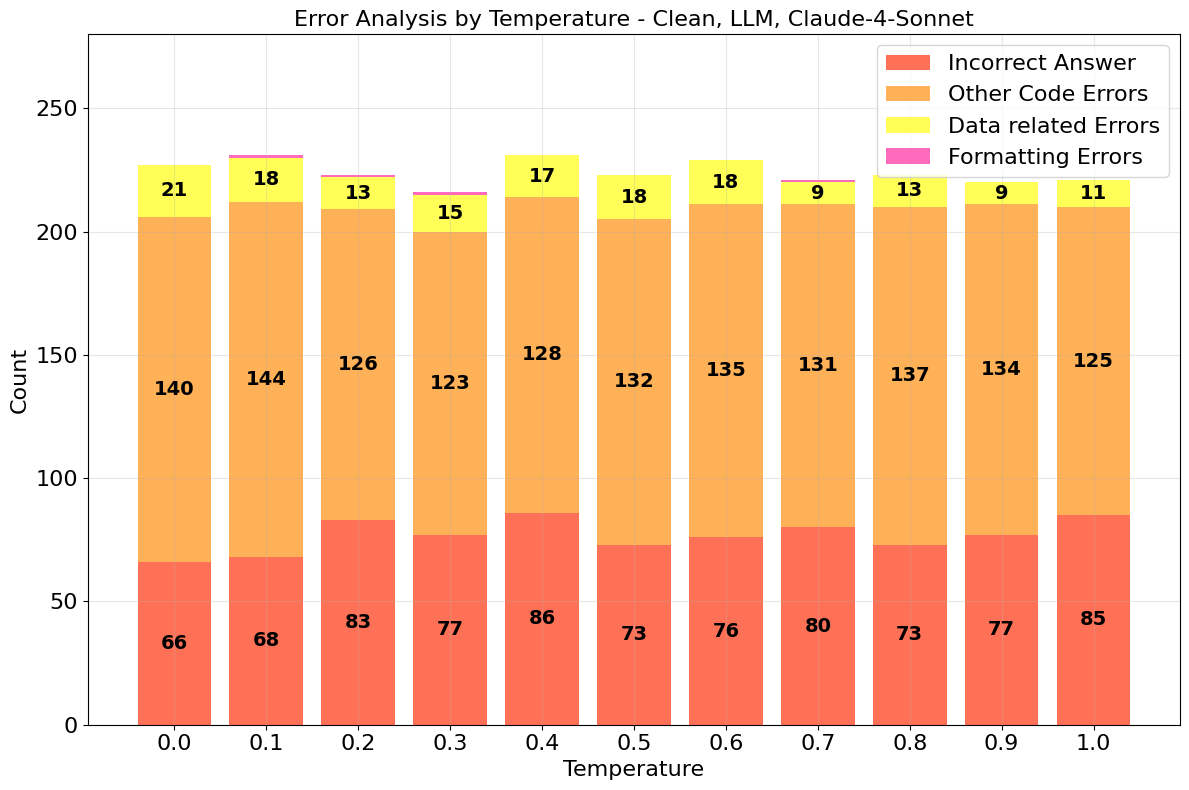}
    \caption{Incorrect response cause distribution - }
    \label{figure:08L}
\end{figure*}
\newpage
\begin{figure*}[!h]
    \centering
    \includegraphics[width=1\linewidth]{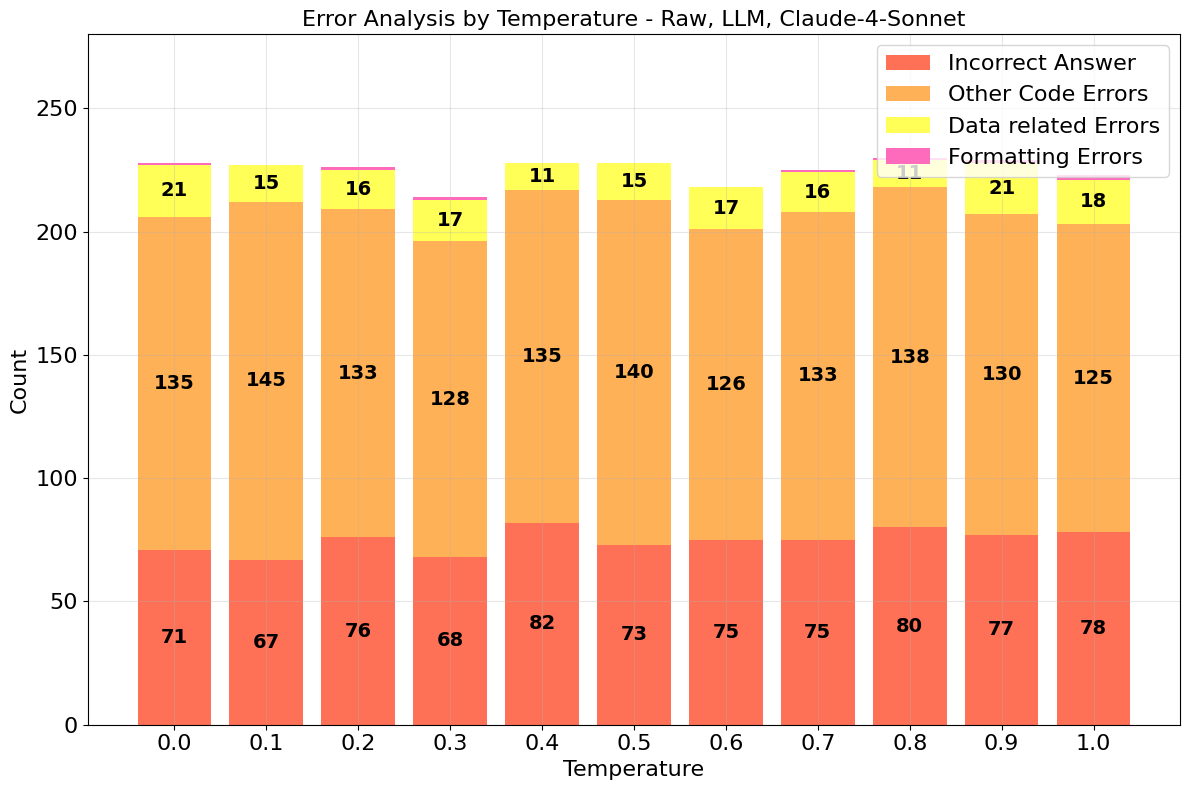}
    \caption{Incorrect response cause distribution - }
    \label{figure:08M}
\end{figure*}
\begin{figure*}[!h]
    \centering
    \includegraphics[width=1\linewidth]{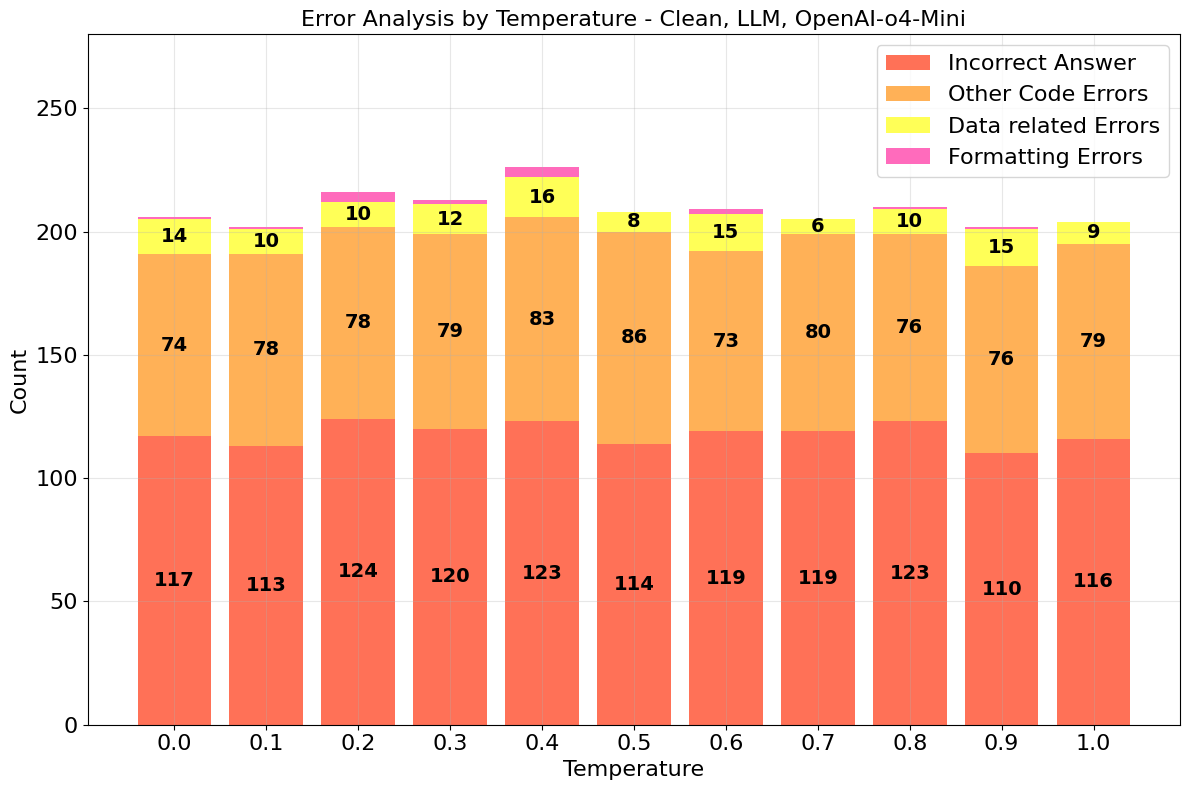}
    \caption{Incorrect response cause distribution - }
    \label{figure:08N}
\end{figure*}
\newpage
\begin{figure*}[!h]
    \centering
    \includegraphics[width=1\linewidth]{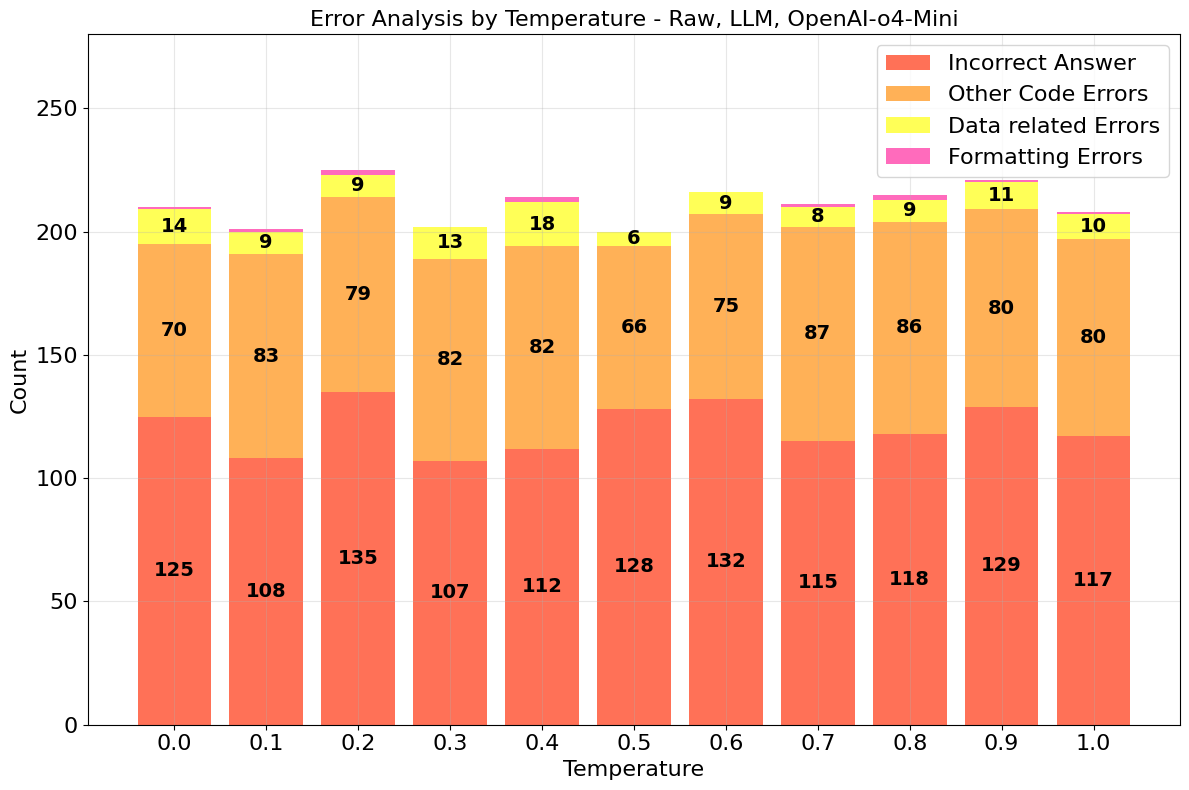}
    \caption{Incorrect response cause distribution - }
    \label{figure:08O}
\end{figure*}
\section{Other Plots}
\label{sec:other}
\begin{figure*}[!h]
    \centering
    \includegraphics[width=1\linewidth]{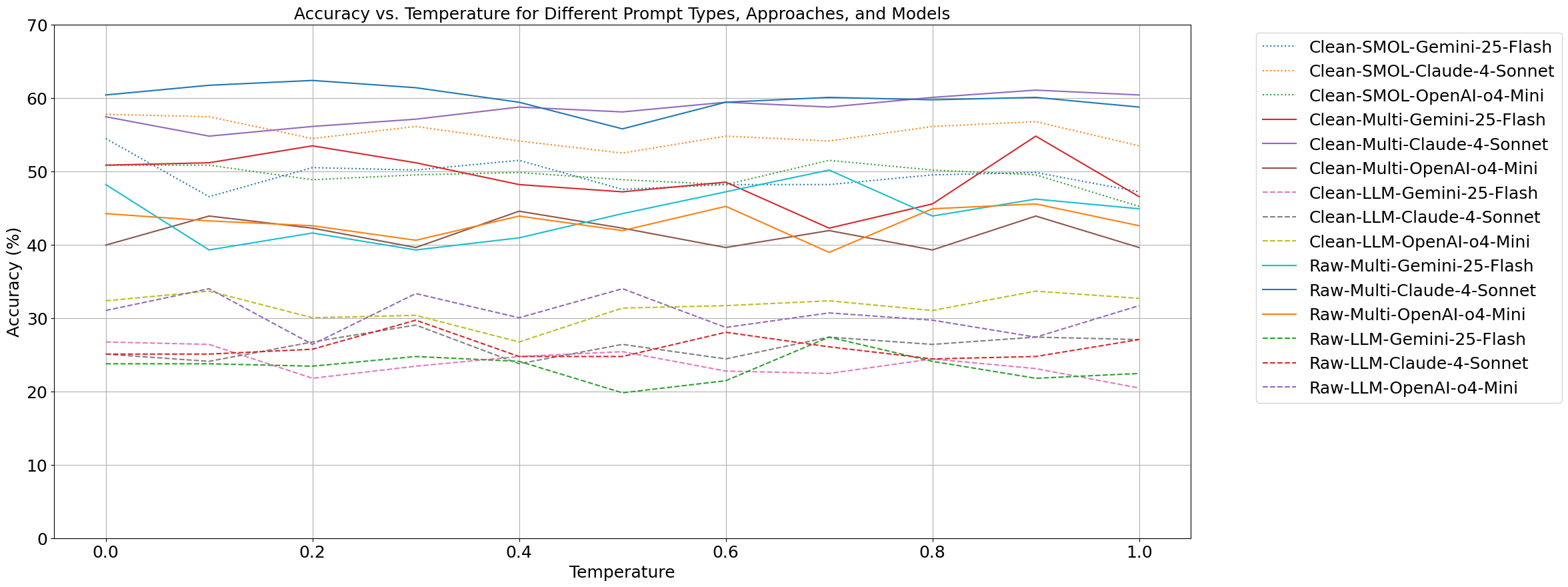}
    \caption{Results from each attempt over each temperature value used}
    \label{figure:2}
\end{figure*}
\newpage
\end{document}